\documentclass{article}

\usepackage{pdfpages}
\usepackage{amsmath}
\usepackage{mathtools}
\usepackage{amsthm}

\usepackage{microtype}
\usepackage{graphicx}
\usepackage{booktabs} % for professional tables
\usepackage{algorithm}
\usepackage{algorithmic}
\usepackage{listings}

\usepackage[backref=page,colorlinks=true,linkcolor=blue,citecolor=green,urlcolor=blue]{hyperref}
\usepackage{url}
\usepackage{graphicx}
\usepackage{bbm}
\usepackage{placeins}
\usepackage{adjustbox}
\usepackage{xcolor}
\usepackage{natbib}
\usepackage[outdir=./]{epstopdf}
\usepackage{graphicx,float,pgfplots,wrapfig,sidecap,lipsum}

\DeclareMathOperator*{\argmin}{arg\,min}
\usepackage{colortbl}
\usepackage{tablefootnote}
\usepackage[font=small,labelfont=bf]{caption}
\usepackage{subcaption}
\usepackage{subfloat}
\usepackage{tikz}
\usetikzlibrary{fit}
\usetikzlibrary{calc,shapes}
\usetikzlibrary{decorations.pathmorphing} % noisy shapes
\usetikzlibrary{fit}					% fitting shapes to coordinates
\usetikzlibrary{backgrounds}	% drawing the background after the foreground
\usetikzlibrary{pgfplots.groupplots}

\usepackage[utf8]{inputenc}
\usepackage{pgfplots}
\usepgfplotslibrary{groupplots,dateplot}
\usetikzlibrary{patterns,shapes.arrows}
\pgfplotsset{compat=newest}

\usepackage{xspace}
\usepackage{soul}

\usepackage[capitalise]{cleveref}

\usepackage{listings}
\usepackage{multirow}
\usepackage{xcolor}
\usepackage{mathrsfs}

\usepackage{enumitem}

\definecolor{codegreen}{rgb}{0,0.6,0}
\definecolor{codegray}{rgb}{0.5,0.5,0.5}
\definecolor{codepurple}{rgb}{0.58,0,0.82}
\definecolor{backcolour}{rgb}{0.95,0.95,0.92}

\lstdefinestyle{mystyle}{
    backgroundcolor=\color{backcolour},   
    commentstyle=\color{codegreen},
    keywordstyle=\color{magenta},
    numberstyle=\tiny\color{codegray},
    stringstyle=\color{codepurple},
    basicstyle=\ttfamily\footnotesize,
    breakatwhitespace=false,         
    breaklines=true,                 
    captionpos=b,                    
    keepspaces=true,                 
    numbers=left,                    
    numbersep=5pt,                  
    showspaces=false,                
    showstringspaces=false,
    showtabs=false,                  
    tabsize=2
}

\usepackage{bm}
\usepackage{makecell}

\theoremstyle{remark}

\AtBeginEnvironment{proof}{}{}{}

\newcommand{\sg}{\text{stopgrad}}

\definecolor{sourcecolor}{rgb}{0.5,1,0.5}
\definecolor{ourcolor}{rgb}{1,0,0}
\definecolor{singlecolor}{rgb}{0,0,1}
\definecolor{auxcolor}{rgb}{0.54,0.17,0.89}
\definecolor{linearcolor}{rgb}{0.172549019607843,0.627450980392157,0.172549019607843}
\definecolor{randomcolor}{rgb}{1,0.498039215686275,0.0549019607843137}
\definecolor{tunecolor}{rgb}{0.9568627450980393, 0.8156862745098039,0}
\definecolor{aligncolor}{rgb}{0,0.5,0}

\usepackage{amsfonts}
\usepackage{amssymb}
\usepackage{xspace}

\newcommand{\algo}{\texttt{PRISE}\xspace}
\usepackage[accepted]{icml2024}
\usepackage{amsmath}
\usepackage{amssymb}
\usepackage{mathtools}
\usepackage{amsthm}
\newcommand{\bpe}{Byte Pair Encoding\xspace}
\usepackage[textsize=tiny]{todonotes}

\icmltitlerunning{PRISE: LLM-Style Sequence Compression for Learning Temporal Action Abstractions in Control}

\begin{document}

\twocolumn[
\icmltitle{PRISE: LLM-Style Sequence Compression for Learning \\ Temporal Action Abstractions in Control}

\icmlsetsymbol{equal}{*}

\begin{icmlauthorlist}
\icmlauthor{Ruijie Zheng}{umd,partmicrosoft}
\icmlauthor{Ching-An Cheng}{microsoft}
\icmlauthor{Hal Daum\'e III}{umd,microsoft}
\icmlauthor{Furong Huang}{umd}
\icmlauthor{Andrey Kolobov}{microsoft}\\
\ \\
\textit{The code is available at:}  \\\url{https://ruijiezheng.com/project/PRISE/index.html}
\end{icmlauthorlist}

\icmlaffiliation{umd}{Department of Computer Science, University of Maryland, College Park}
\icmlaffiliation{partmicrosoft}{Part of the work done while at Microsoft}
\icmlaffiliation{microsoft}{Microsoft Research}
\icmlcorrespondingauthor{Ruijie Zheng}{rzheng12@umd.edu}

\icmlkeywords{Machine Learning, ICML}

\vskip 0.3in
]

\printAffiliationsAndNotice{}  

\begin{abstract}
Temporal action abstractions, along with belief state representations, are a powerful knowledge sharing mechanism for sequential decision making.
In this work, we propose a novel view that treats inducing temporal action abstractions as a sequence compression problem. To do so, we bring a subtle but critical component of LLM training pipelines -- input tokenization via byte pair encoding (BPE) -- to bear on 
the seemingly distant task of learning skills of variable time span in continuous control domains. We introduce an approach called Primitive Sequence Encoding (\algo) that combines continuous action quantization with BPE to learn powerful action abstractions. We empirically show that high-level skills discovered by \algo from a multitask set of robotic manipulation demonstrations significantly boost the performance of Behavior Cloning on downstream tasks.
\end{abstract}
\section{Introduction}

High-dimensional observations and decision-making over complex, continuous action spaces are hallmarks of typical scenarios in sequential decision-making, including robotics. A common way of dealing with these complexities is constructing \emph{abstractions} -- compact representations of belief states and actions that generalize across tasks and make learning to act in new scenarios robust and data-efficient. Inspired by techniques that have enabled ground-breaking progresses in computer vision (CV) and natural language processing (NLP) over the past decade, much of continuous control research has focused on \emph{learning} abstractions from data rather than hand-crafting them. The lion's share of these efforts study learning multi-task representations, which has analogies to representation learning in CV and NLP and has been tackled by adapting these fields' models (see, e.g., DT~\citep{chen2021decision} vs. GPT-2~\citep{radford2019GPT2}) and methods (see, e.g., R3M~\citep{nair2022r3m} vs. InfoNCE~\citep{oord2019representation, zheng2023taco}). 

On the other hand, learning \emph{temporal action abstractions} -- representations of multi-step primitive behaviors  -- has not benefited nearly as much from such a methodology transfer. This omission is especially glaring in continuous control, where complex policies can clearly be decomposed into versatile lower-level routines such as picking up objects, walking, etc, and whose popular learning method, Behavior Cloning (BC)~\citep{arora2023exposure}, has many commonalities with LLM training.

In this paper, we posit that adapting  discrete coding and sequence compression techniques from NLP offers untapped potential for action representation learning for continuous control. Specifically, given a pretraining dataset of demonstrations from multiple decision tasks over a continuous action space and a high-dimensional pixel observation space, we consider the problem of learning temporally extended action primitives, i.e., \emph{skills}, to improve downstream learning efficiency of reward-free BC. We show that embedding continuous actions into \emph{discrete} codes and then applying a popular NLP sequence compression method called \emph{\bpe(BPE)}~\citep{gage94bpe} to the resulting discrete-code sequences identifies variable-timespan action primitives with the property we desire. Namely, in downstream tasks, policies learned by BC using these action primitives \emph{consistently perform better}, often substantially, than policies learned by BC directly over the original action space. 

Our work's main contribution is \emph{\textbf{Pri}mitive \textbf{S}equence \textbf{E}ncoding (\algo)}, a novel method for learning multi-task temporal action abstractions that capitalizes on a novel connection to NLP methodology. \algo quantizes the agent's original continuous action space into discrete codes, converts the pretraining training trajectories into action code sequences, and uses BPE to induce variable-timestep skills. During BC for downstream tasks, learning policies over these skills and then decoding skills into primitive action sequences gives \algo a significant boost in learning efficiency over strong baselines such as ACT~\citep{zhang2021learning}. We conduct extensive ablation studies to show the effect of various parameters on \algo's performance, which demonstrate BPE to be critical to \algo's success. 
\section{Preliminaries\label{sec:prelims}}
\paragraph{Problem Setting.} We consider a set of tasks $T$, denoted as $\big\{(\mathcal{O}, \mathcal{A}, \mathcal{P}, g_{\mathcal{T}})\big\}_{\mathcal{T} \in \mathscr{T}}$, that share the action space $\mathcal{A}$, the transition dynamics $\mathcal{P}$, and the observation space $\mathcal{O}$. The  tasks differ only in their goals $g_{\mathcal{T}}$.
Since we focus on learning visuomotor continuous control policies, we assume $\mathcal{A}$ is a finite-dimensional vector space and $\mathcal{O}$ is the image space. \looseness=-1

Let $\mathcal{D}=\bigcup_{\mathcal{T} \in \mathscr{T}}\mathcal{D}_{\mathcal{T}}$ 
represent a multitask dataset, with each $\mathcal{D}_{\mathcal{T}}$ containing expert trajectories specific to task $\mathcal{T}$.
The focus of our work is to leverage the multi-task dataset $\mathcal{D}$ to learn a vocabulary $\mathcal{V}$ of skill \emph{tokens} representing temporally extended low-level policies, which we call \emph{skills}. Note that our concept of a skill is different from some others in prior work; we compare them in \cref{sec:main:relw}. Here skill policies can be thought of as control primitives for behaviors that are common across many tasks, e.g., lowering a robotic arm's end-effector to the table-top level or flipping an object the end-effector is holding. Due to their universality, these primitives can enable efficient few-shot imitation learning (IL) for unseen tasks.
Formally, we define the skill token vocabulary as $\mathcal{V} = \{\xi_1, \ldots, \xi_K\}$ , where each skill token $\xi$ is a tuple $\langle f_\xi,L_\xi \rangle$ s.t. if the agent chooses token $\xi$ at time step $t$, then for the next $L_\xi$ steps it will take actions $a_{t+k}=f_\xi(o_{t+k}, k)$, $0 \leq k <  L$ recommended by the token's skill policy $f_\xi$.

\vspace{0.2em}
\textbf{Vector Quantization.} In order to construct the temporal action abstraction tokens, our algorithm \algo~ first converts continuous actions into discrete codes. To do so, it uses the vector quantization module proposed in~\citet{vqvae}. This vector quantization module $\mathcal{F}$ maintains a collection $\mathcal{E}=\{e_{1},..., e_{C}\}$ of $C$ codes. For any query vector $q$ (in our case, $q$ is a concatenation of latent state and action, as described in \Cref{sec:pretr1}), the module maps $q$ to $\mathcal{F}(q)=e_k$, where $e_k=\argmin_{e_i \in \mathcal{E}}\|e_i-q\|$. During training, $\mathcal{F}$ is optimized w.r.t. the loss  $\mathcal{L}[\theta_{\mathcal{F}}] = \|\sg(q)-e_k\|_2+\|q-\sg(e_k)\|_2$, where $\sg(.)$ stands for stop-gradient. 
The first term here brings the code closer to the query vector $q$, and the second term prevents the query vector from switching between different codes. 

\textbf{\bpe (BPE).} After quantizing the action space into codes, \algo\ identifies common code (action) sequences via a technique from  the realm of NLP, the method of \bpe(BPE)~\citep{gage94bpe, sennrich2016neural}. 
In NLP, BPE has emerged as a pivotal technique for managing the vast and varied vocabulary encountered in textual data. Language models are expected to predict a text completion based on a text prefix. Doing so naively, by generating the text character-by-character is problematic, because the notion of a character differs significantly across languages. Instead, NLP models operate at the language-agnostic level of \emph{bytes}. However, the high granularity of byte prediction has its own computational challenges. Therefore, language models models first find high-frequency byte \emph{sequences} in the training data, assign \emph{tokens} to these sequences, learn to predict the resulting tokens, which then get decoded into text. BPE is the algorithm that handles the common byte sequence construction problem.

\begin{figure}[!htbp]
    \includegraphics[width=\linewidth]{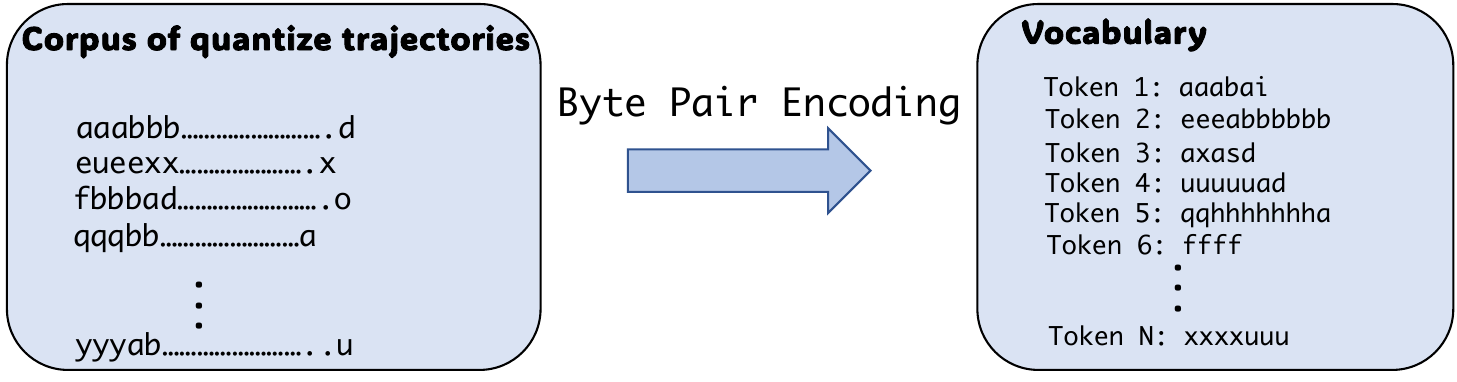}
    \vspace{-2em}
    \caption{Byte Pair Encoding.}\label{fig:BPE}
\end{figure}

BPE operates by merging the most frequent pairs of bytes, assigning tokens to these pairs, and iteratively merging new pairs of tokens or bytes. Thereby, it builds up a vocabulary of tokens that plays a crucial role in modern large language models~\cite{gpt4, brown2020gpt3,radford2019GPT2}. See \Cref{fig:BPE} for a demonstration. We apply BPE in the same way, but to action codes instead of bytes.

\section{Algorithm}
\label{s3:algorithm}

\begin{figure*}[ht!]
\centering
\includegraphics[width=1.0\linewidth]{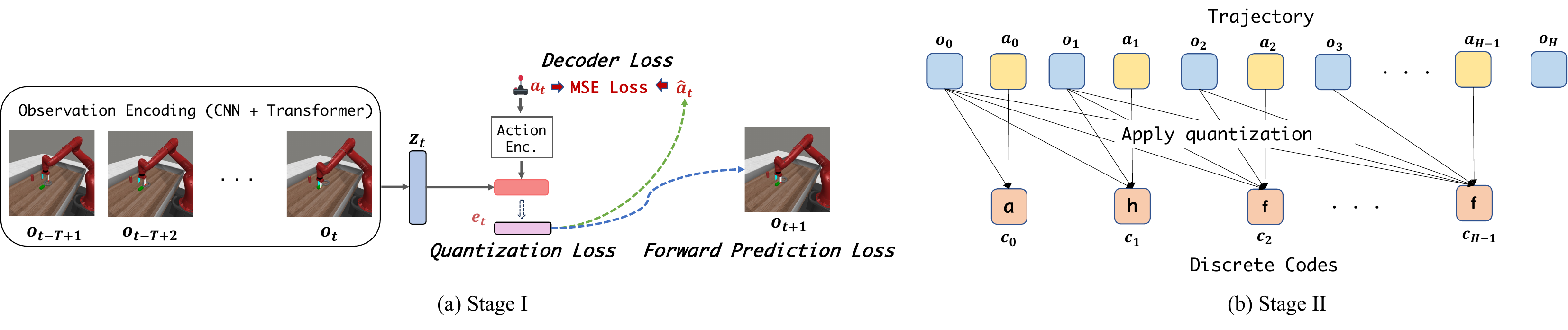}
\vspace{-2.5em}
\caption{(\textbf{a}) Pretraining Stage I of \algo~: The goal is to learn a action quantization module such that conditioned on the current state and action $(o_t,a_t)$, it could assign a discrete action code. 
(\textbf{b}) Pretraining Stage II of \algo~: First it converts a trajectory of continuous state and actions into discrete codes. Then based on the corpus of quantized trajectories from the multitask offline dataset, \algo~applies BPE (illustrated in \Cref{fig:BPE}) to learn vocabulary of skill tokens, where each token represents a sequence of discrete action code.
}
\label{fig: pretrain_diagram}
\vspace{-0.5em}
\end{figure*}

\Cref{fig: pretrain_diagram} illustrates the key steps of \algo.
At a high level, \algo first learns a state-dependent action quantization module (Pretraining I stage). This module processes the pretraining multitask dataset $\mathcal{D}$ by transforming each of its trajectories -- a sequence of $\langle$observation, continuous action$\rangle$ pairs -- into a sequence of discrete codes, one code per time step, as shown in \Cref{fig: pretrain_diagram}(a). Then, in Pretraining II stage, \algo employs the BPE tokenization algorithm to identify commonly occurring code sequences in the processed dataset $\mathcal{D}$, assigns a token to each of them, and thereby constructs primitive skills, shown in \Cref{fig: pretrain_diagram}(b).
Once these primitive skill tokens are discovered, they can be used to relabel a demonstration dataset and learn via behavior cloning (BC) a policy that chooses these skill tokens instead of raw actions, as we later show in \Cref{sec:multitask policy learning} and \Cref{sec:adaptation}.
In the rest of the section, we describe \algo in more detail. 

\subsection{Pretraining \label{sec:pretr1}}
\textbf{Pretraining I: action quantization}. This stage is illustrated in \Cref{fig: pretrain_diagram}(a), and its pseudocode is available in \Cref{fig:pseudocode} in \Cref{a2:implementation}. Let $\mathcal{G}: \mathcal{O}^T \rightarrow \mathcal{Z}$ be the observation embedding which embeds a sequence of pixel observations of length $T$ into the latent embedding space $\mathcal{Z}$. 
In our implementation, we use the transformer architecture ~\citep{vaswani17att} and set $z_t = \mathcal{G}(o_t,o_{t-1},..., o_{t-T+1})=\textit{Transformer\_block}(\text{CNN}(o_t), ,..., \text{CNN}(o_{t-T+1}))$. Using the multitask dataset $\mathcal{D}$, we jointly pretrain the observation embedding $\mathcal{G}$ and a state-dependent action quantization module $\mathcal{F}: \mathcal{Z}\times \mathcal{A}\rightarrow \mathcal{E}$, where $\mathcal{E}=\{e_{1},e_{2},...,e_{C}\}$ denotes the set of codes. 
The set of codes $\mathcal{E}$ produced by learning $\mathcal{F}$ depends on the inductive bias of $\mathcal{F}$'s training procedure. The inductive bias we chose for \algo\ follows the intuition that, given the current latent state, a desirable action code should be predictive of both the next latent state and of the raw action mapped by $\mathcal{F}$ into that action code. A similar intuition underlies action representation learning in, e.g., \citep{chandak2019learning}, although that work doesn't attempt to simultaneously learn a latent state space.

First, to ensure the action code can predict future states, we 
train a latent forward transition model $\mathcal{T}: \mathcal{Z}\times \mathcal{A}\rightarrow \mathcal{Z}$. 
To optimize the forward latent dynamics model while preventing state and action representation collapse, we adopt a BYOL-like objective inspired from~\citep{schwarzer2021pretraining}, where we minimize
\begin{align}
\mathcal{L}_\text{dyn}[&\theta_{\mathcal{F}, \mathcal{G}, Q, P}](o_{t,T}, a_t)= \\
&-\textbf{cos\_sim}\big(Q(P(\mathcal{T}(z_t,e_t))), \:\sg(P(z_{t+1}))\big) \nonumber
\end{align}
Here, $o_{t,T} = o_t,o_{t-1},..., o_{t-T+1}$ is the observation history, $z_t = \mathcal{G}(o_t,o_{t-1},..., o_{t-T+1})$ is the latent state, $e_t=\mathcal{F}(\sg(z_t),a_t)$ is the quantized action code, $P$ is a projection MLP, and $Q$ is a prediction MLP. In our implementation, $P$ has two layers and $Q$ has one.  

Next, to guarantee the raw action can be decoded from the action code and latent state effectively, we train a latent state-dependent decoder, $\psi$, aiming to reconstruct the raw action given latent state $z_t$ and action code $e_t$.  
In general, we let $\psi$ be a stochastic mapping $\psi: \mathcal{Z}\times \mathcal{E} \rightarrow \Delta(\mathcal{A})$, where $\Delta(\mathcal{A})$ stands for the space of probability distributions over $\mathcal{A}$.
We use a stochastic Gaussian Mixture Model (GMM) to parameterize the decoded action distribution of $\psi$ following~\citet{robomimic2021}. This choice has been found  effective in dealing with the inherent multimodality and noise in such human teleoperation demonstrations.
Then the learning objective becomes to minimize the negative log likelihood of the GMM distribution: $\mathcal{L}_\text{act\_decode}[\theta_{\mathcal{F}, \mathcal{G}, \psi}](o_{t,T}, a_t)=\mathcal{L}_\text{act}(\psi(z_t, e_t), a_t)=-\log \psi(a_t|z_t,e_t)$.
When we know the the pretraining data is collected by a fixed deterministic policy, we let $\psi: \mathcal{Z}\times\mathcal{E} \rightarrow \mathcal{A}$ be a deterministic function that minimizes the $L1$-loss:
$\mathcal{L}_\text{act\_decode}[\theta_{\mathcal{F}, \mathcal{G}, \psi}](o_{t,T}, a_t)= \mathcal{L}_\text{act}(\psi(z_t, e_t), a_t)=\|\psi(z_t, e_t)-a_t\|_1.$

Combining the two objectives, we pretrain the state encoder and action encoder/quantization module by 
\vspace{-0.3em}
\begin{equation}
\label{eq:quant_loss}
\resizebox{0.85\columnwidth}{!}{$\mathcal{L}[\theta_{\mathcal{F}, \mathcal{G}, Q, P, \psi}]=\mathcal{L}_\text{dyn}[\theta_{\mathcal{F}, \mathcal{G}, Q, P}]+\beta\mathcal{L}_\text{act\_decode}[\theta_{\mathcal{F}, \mathcal{G}, \psi}]$}
\end{equation}
Throughout the experiment, we set $\beta=1$ when $\psi$ is a deterministic decoder, and $\beta=0.01$ when $\psi$ is parametrized by GMM to deal with numerical scale of the likelihood loss.

\textbf{Pretraining II: temporal action abstractions via BPE.} \Cref{fig: pretrain_diagram}(b) illustrates the mechanics of this stage. After training the action quantizer, we first use the pretrained observation embedding $\mathcal{G}$, action quantizer $\mathcal{F}$ to convert the  trajectories from pretrainig dataset $\mathcal{D}$ into sequences of discrete quantized action codes, $\big\{(e^{(i)}_0, e^{(i)}_1, ..., e^{(i)}_{H_i})\big\}_{i=1}^{|\mathcal{D}|}$, where $e^{(i)}_t \in \mathcal{E}$ for all $t, i$ and $H_i$ stands for the length of episode $i$. 
Then, drawing analogy to NLP, we view  the sequence of action codes in each trajectory as a ``sentence" and  employ the BPE algorithm to create a vocabulary of skill tokens $\mathcal{V}=\{\xi_1,..., \xi_K\}$, where tokens represent ``sentence''/trajectory segments. To do so, we initialize BPE using codes from $\mathcal{E}$ and let BPE construct the vocabulary $\mathcal{V}$ by iteratively merging the most frequent pairs of action codes or action code sequences into a single token. Different resulting tokens $\xi \in \mathcal{V}$ may correspond to skills with different horizons $L_{\xi}$. 

\textbf{Summary.} Overall, the pretraining of \algo produces observation embedding $\mathcal{G}$, action quantizer $\mathcal{F}$, and a vocabulary $\mathcal{V}$ of skill tokens. In the following, we discuss how to use them to learn generalist multitask policies and achieve efficient downstream adaptation.

\subsection{Multitask generalist policy learning.} \label{sec:multitask policy learning}
Once we have learned the skill tokens that capture the common motion patterns shared across various tasks, we can leverage these skills to learn a multitask generalist policy.
We train a high-level skill-token policy $\pi: \mathcal{Z} \rightarrow \Delta(\mathcal{V})$ that predicts a skill token based on a latent state. 
At rollout time, $\pi$ works as follows.
At timestep $t$, suppose $\pi$ predicts a token $\xi_t$ after seeing observation $o_{t,T}$ and the token $\xi_t$ corresponds to $L_{\xi_t}$ action codes, without loss of generality, $e_{t+k}$, $0 \leq k < L_{\xi_t}$.
Then for the subsequent $L_{\xi_t}$ steps, the agent applies the codes $e_{t+k}$, $0 \leq k < L_{\xi_t}$ in their order, and uses the decoder $\psi: Z\times \mathcal{E}$  learned in Pretraining I stage to convert each $e_{t+k}$ to an action $a_{t+k}$ from the original action space $\mathcal{A}$. 
In other words, $a_{t+k}\sim \psi(z_{t+k}, e_{t + k})$ for each $k$ s.t. $0\leq k < L_{\xi_t}$. After $L_{\xi_t}$ steps, the agent queries $\pi$ again to choose the next sequence of actions.
\begin{figure}[h!]
\centering
\vspace{-0.5em}
\includegraphics[width=1.0\columnwidth]{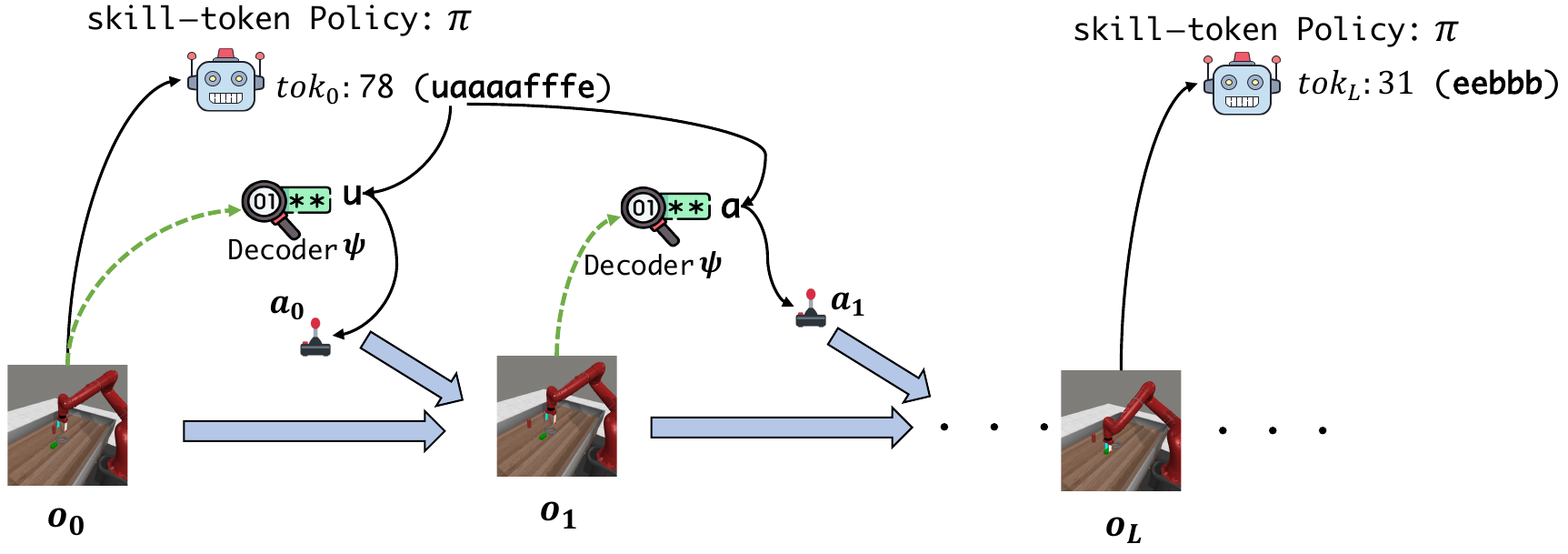}
\vspace{-2.0em}
\caption{During evaluation time, \algo~rollout its policy by querying the skill-token policy $\pi$ for the skill token and then using pretrained decoder $\psi$ to decode actions.}
\vspace{-1.0em}
\label{fig: policy_rollout_seq}
\end{figure}

To learn this skill-token policy $\pi$ from a set of expert demonstration trajectories $\mathcal{D} = \{(o^{(i)}_0,a^{(i)}_0,...o^{(i)}_{H_i})\}_{i=1}^{|\mathcal{D}|}$, we first run the action quantizer $\mathcal{F}$ and observation encoder $\mathcal{G}$ to get the quantized action codes of the trajectories $\{(c^{(i)}_0, c^{(i)}_1, ...o^{(i)}_{H_i})\}_i$, as well as latent observation embeddings $\{(z^{(i)}_0, z^{(i)}_1, ...z^{(i)}_{H_i})\}_i$.
Then for each timestep $t$ in episode $i$, we compute the target expert token $\xi^{(i)}_t$ by greedily searching for the longest token in the BPE tokenizer's vocabulary that matches the sequence $(c^{(i)}_t, c^{(i)}_{t+1},..., c^{(i)}_{H_i})$. 
\begin{figure}[h!]
\centering
\vspace{-0.5em}
\includegraphics[width=0.85\columnwidth]{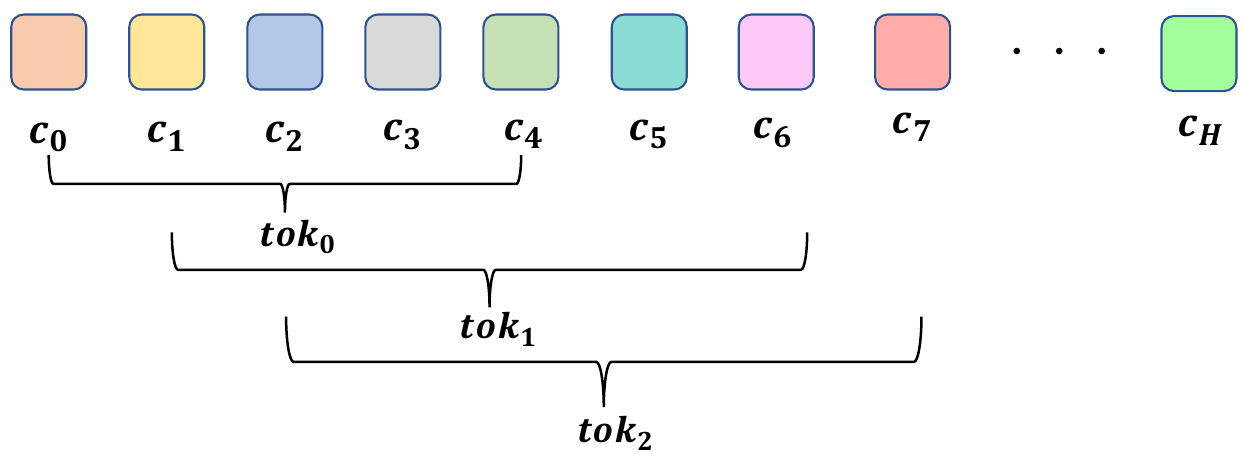}
\vspace{-1.5em}
\caption{\algo~tokenizes downstream demonstration trajectories by greedily searching for the longest token for each time step.}
\vspace{-0.5em}
\label{fig: tokenization}
\end{figure}
\newline

We then train $\pi$ by minimizing the cross-entropy loss: 
\begin{equation}
\mathcal{L}_{\textbf{CE}}(\pi(\sg(z_t)), \xi_t)[\theta_{\pi}]
\label{eq:ce}
\end{equation}
Note, that as $\sg$ implies, we freeze the encoder $\mathcal{G}$. 
\looseness=-1

\subsection{Downstream few-shot adaptation to unseen tasks} \label{sec:adaptation}
In addition to learning a generalist multitask policy $\pi$, we can also use the pretrained skill tokens to adapt $\pi$ to unseen downstream tasks with a few expert demonstrations.
Unlike in the multitask pretraining setting, during finetuning we need to finetune the decoder $\psi$ for downstream unseen tasks, in addition to adapting $\pi$ itself.

We optimize $\psi$ with the following objective, where the summation over training trajectories and time steps $t$ in each trajectory has been omitted for clarity:
\vspace{-0.3em}
\begin{equation}
\resizebox{0.8\columnwidth}{!}%{$\mathcal{L}_{\textbf{FT\_DECODER}}(\pi, \psi)=\mathbb{E}_{\xi\sim \pi(\sg(z_i))}\Big[\mathcal{L}(\psi, \xi)\Big]$}
{$\mathcal{L}_{\textbf{FT\_DECODER}}[\theta_{\pi, \psi}]=\mathbb{E}_{\xi_t\sim \pi(\sg(z_t))}\Big[\mathcal{L}[\theta_{\psi}](\xi_{t})\Big]$}
\end{equation}
\vspace{-0.5em}
where
\begin{equation}
\resizebox{0.8\columnwidth}{!}
{$\displaystyle \mathcal{L}[\theta_{\psi}](\xi_t)=\sum_{i=0}^{\hat{K}}\mathcal{L}_{\text{action}}(\psi(\sg(z_{t+i}), \xi_t[i]), a_{t+i})$}
\end{equation}
In this equation, $\hat{K}=\min(K, L_{\xi})$, where $K$ is a hyperparameter, the motivation behind which is explained at the end of \cref{sec:discussion of HPs}. 
$\xi_t[i]$ represents the $i$-th code of the skill token $\xi_t$.

We optimize skill-token policy $\pi$ w.r.t. the cross-entropy loss $\mathcal{L}_\textbf{CE}$ in \eqref{eq:ce}.
Thus, the overall objective is 
\vspace{-0.3em}
\begin{equation}
\resizebox{0.95\columnwidth}{!}{$\displaystyle \mathcal{L}[\theta_{\pi, \psi}] = \mathcal{L}_\textbf{CE}(\pi(\sg(z_t)), \xi_t) + \mathcal{L}_{\textbf{FT\_DECODER}}[\theta_{\pi, \psi}]$}
\label{eq:dstream_overall}
\end{equation}
\vspace{-1.5em}

Note that the gradients of the decoder finetuning loss $\mathcal{L}_{\textbf{FT\_DECODER}}$ not only backpropagate into the decoder $\psi$ but also into the skill-token policy $\pi$ so that $\pi$ should be aware of the decoder error.

The underlying rationale for making the objective pay attention to matching both the skill tokens and the actions that these tokens generate is as follows. The number of skill tokens in the vocabulary is often large in order to encapsulate all motion patterns observed in the pretraining dataset, and simply minimizing the cross-entropy loss over the tokens with sparse data from few expert trajectories does not suffice to learn an accurate skill-token policy. 
Instead, the objective in \eqref{eq:dstream_overall} says that the skill-token policy $\pi$ should not only match the target token but also predict the tokens that have small decoding errors.

\subsection{Discussion} \label{sec:discussion of HPs}

\algo combines state-action abstraction (by optimizing the quantization loss in \Cref{eq:quant_loss}) and a temporal flavor of action abstraction (by applying BPE tokenization) to reduce the complexity of downstream IL problems. It is known that the complexity of IL depends on mainly two factors: the size of the state-action space and the problem horizon~\citep{rajaraman2020toward}. The former determines the minimum complexity of the learner's policy. The latter determines how much the behavior cloning error will compound. 
\algo addresses these two factors by its state-action abstraction and temporal abstraction, respectively.

The effectiveness of \algo's abstraction scheme is determined by several hyperparameters.
First, \textbf{the codebook size}, which determines the granularity of quantization, trades off the approximation error of state-action abstraction and the complexity of the state-action representation that the learner's policy sees. 
We want the codebook to be large enough to represent the demonstrator's policy well, but not too large in order to avoid several codes collapsing to the same set of actions in the original space.  
Many duplicate codes would reduce the amount of temporal abstraction \algo can achieve.

Second, \textbf{the size of the token vocabulary}, which is at least as large as the codebook size by the design of BPE, not only affects the complexity of the state-action representation, but also controls how generalizable the learned tokens are. 
When the token vocabulary is large, BPE picks up less frequent patterns, which typically are longer. Although these tokens may help compress the length of the training data, they might overfit to the training trajectories and  end up never getting used in downstream tasks. Moreover, the overfit tokens might be too long and lead to approximation errors when decoded back to the original action space. \algo's hierarchical learning scheme implicitly assumes the expert policy applies the action codes in an \emph{open-loop} manner up the horizon $L_{\xi}$ of the skill token $\xi$ that is decoded into these codes. This assumption stems from how the policy is used in roll out (see \cref{fig: policy_rollout_seq}), and is related to also the Pretraining II stage, where BPE assembles skill tokens from action code sequences in an open-loop fashion ignoring latent states. In practice, the demonstrator's policy may be closed-loop, in which case our approximation of it with open-loop skill policies introduces an error. This error grows with the length of a skill's open-loop action sequence. Therefore, we introduce a hyperparameter $K$ (see \cref{sec:adaptation}) that caps skills' possible horizon in downstream adaptation.

On the contrary, when the token vocabulary size is set to be too small, the effects of temporal abstraction are minimal and would not reduce learning complexity. 
In the extreme, when the vocabulary size is the same as the codebook size, \algo would only perform state-action abstraction by learning a discrete action space with state-action dependent encoder and decoder, without reducing the compounding error.\looseness=-1

\section{Related Work } \label{sec:main:relw}

Please refer to \Cref{sec:appendix:relw} for an extensive discussion of related work. Here we provide its summary

\textbf{Temporal Action Abstractions in Sequential Decision-Making}: Temporal action abstraction and options or skills by compression is not newly introduced in this paper. In fact, the concept of temporal action abstraction is rooted in the literature of both fully and partially observable MDPs \citep{puterman94, thrun94, sutton2018}. The complexity of policy learning is proportional to the problem horizon \citep{ross2011reduction,jiang2015,rajaraman2020toward,xie2022bellmanconsistent}, leading to hierarchical approaches \citep{parr98rl,barto03advances,le2018hierarchical,nachum2018dataefficient,kipf2019compile,kujanpaa23hierquantized, jiang2022learning} and the use of options or primitives \citep{sutton99options,ajay2021opal, cui2023cbet}.

\textbf{Temporal Abstraction Discovery for Decision-Making}: Various methods like CompILE \citep{kipf2019compile}, RPL \citep{gupta20relay}, OPAL \citep{ajay2021opal}, TAP \citep{jiang2023efficient}, and ACT \citep{aloha} learn temporally extended action primitives in two stages. DDCO \citep{krishnan17ddco}, in contrast, learns both primitives and high-level policy simultaneously.

 \algo differs in its use of BPE for inducing temporal action abstractions, avoiding challenges faced by CompILE \citep{kipf2019compile}, RPL, and OPAL. ACT \citep{zhao2023learning} is the closest to \algo, especially in using BC, not RL, for downstream learning and handling pixel observations rather than low-level states.

\textbf{Policy Quantization Methods}: \algo relates to SAQ \citep{luo2023actionquantized} and employs VQ-VAE \citep{oord2018neural}, with Conditional VAE \citep{kingma2014vae} being another common choice.

\textbf{Temporal Abstractions vs. Skills in Robot Learning}: Distinct from skill acquisition approaches such as LOVE~\citep{jiang2022learning}, BeT~\citep{Shafiullah2022bet}, C-BeT~\citep{cui2023cbet}, DMP \citep{pastor09skills}, Play-LMP \citep{lynch2019learning}, and MCIL \citep{mcil21rss}, \algo and methods like TAP \citep{jiang2023efficient} focus on learning temporally extended action primitives.

\textbf{Pretraining Data Assumptions}: \algo aligns more with C-BeT \citep{cui2023cbet}, Play-LMP \citep{lynch2019learning}, RPL \citep{gupta20relay}, MCIL \citep{mcil21rss}, and TACO-RL \citep{rosete2022tacorl}, requiring meaningful behavioral patterns in pretraining data.

\textbf{Tokenization in Language Models}: BPE \citep{gage94bpe}, Unigram \citep{kudo2018subword}, and WordPiece \citep{devlin2018bert} are crucial in training language models. \algo extends the next-token-prediction paradigm to continuous control, dealing with challenges like high-dimensional image observations and continuous action spaces.

\section{Experiments}

In this section, we empirically evaluate the effectiveness of the skill tokens of \algo. We run \algo to pretrain these tokens using large-scale, multitask offline datasets. Then we evaluate them on two offline IL scenarios: learning a multitask generalist policy and few-shot adaptation to unseen tasks.
We show that \algo is able to improve the performance of the learner policy, both compared to an \algo version approaches that doesn't use the skill tokens and compared to a very strong existing approach, ACT~\cite{zhao2023learning}.

\subsection{Pretraining datasets and architecture}
We evaluate \algo~on two multitask robotic manipulation benchmarks: Metaworld~\citep{metaworld} and LIBERO~\citep{libero}.
Below we describe the setups of these datasets and the architectures used in the experiments.

\begin{figure}[!htbp]
\centering
\vspace{-0.3em}
\includegraphics[width=\linewidth]{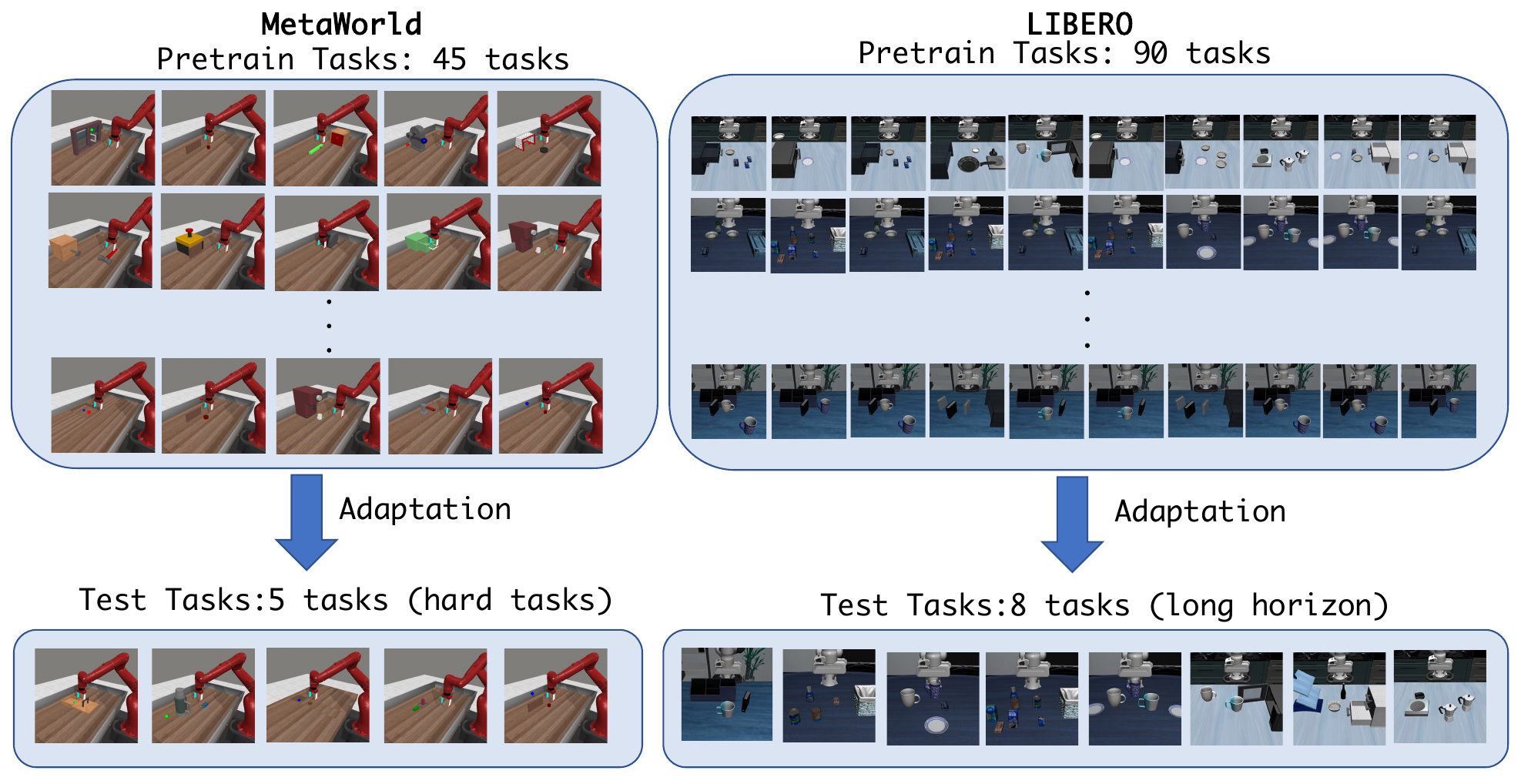}
\vspace{-2em}
\caption{Few-shot IL on unseen tasks: pretrain and test tasks split for MetaWorld and LIBERO.}
\label{fig: pretrain tasks}
\end{figure}

For LIBERO, we pretrain on the 90 short-horizon tasks (LIBERO-90) with offline dataset provided by the original paper. 
We test the learned skill tokens both in terms of the multitask learning performance on LIBERO-90 as well as 5-shot imitation learning (IL) performance on the first 8 unseen tasks from LIBERO-LONG.
The pretraining dataset contains 50 demonstration trajectories for each task, collected by human teleoperation.
We use the exact architecture of ResNet-T in~\cite{libero} for the observation embedding. 

For MetaWorld, we focus on 5-shot IL since multitask IL on pretraining tasks is straightforward, with baseline algorithms, including \algo, all achieving an average success rate of around 80\%. We refer the readers to~\Cref{fig:metaworld_mt45} in~\Cref{a3:multitask} for the results.
For the pretrain-test split, we hold out five hard tasks (hand-insert, box-close, disassemble, stick-pull, pick-place-wall) for few-shot evaluation and perform pretraining on the rest 45 tasks. 
We generate 100 expert trajectories for each pretraining task using the scripted policy provided in MetaWorld. 
Same as in as in~\cite{yarats2022drqv2, xu2024drm}, we use a shallow CNN encoder to encode the observation, and a transformer decoder module as the temporal backbone to encode temporal information into the latent embedding $z_t$.

For BPE on both domains, we set the codebook size $C$ to be 10 and vocabulary size to be 200, and we provide detailed ablation study to analyze their impacts later in this section. 
For more implementation details, we refer the readers to Appendix~\ref{a2:implementation}.

\begin{figure}[!t]
\centering
\includegraphics[width=0.85\columnwidth]{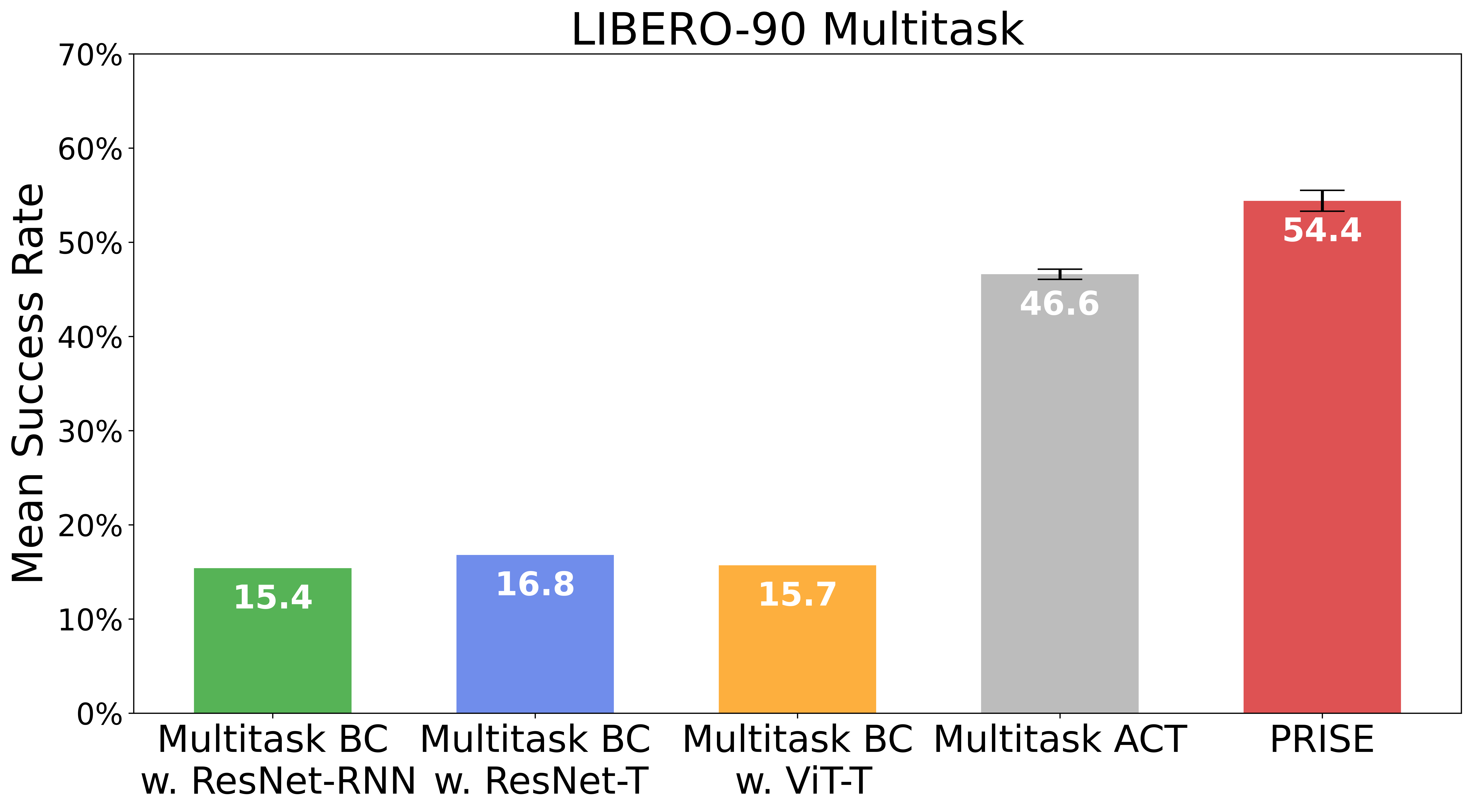}
\vspace{-1em}
\caption{Multitask policy learning performance on \textbf{LIBERO-90.} Error bar represents the standard deviation across 4 random seeds. Results of the first three methods are taken directly from~\citep{libero}}
\vspace{-1.5em}
\label{fig:libero90_multitask}
\end{figure}

\subsection{Multitask generalist policy learning}

First we evaluate whether the pretrained skill tokens of \algo could enable efficient knowledge sharing across tasks and improve multitask IL performance.
Here we focus on \textbf{LIBERO-90}, a challenging multi-task benchmark where existing algorithms and architectures have not demonstrated satisfactory performance. 
For each algorithm in the following comparison, we first perform pretraining on the LIBERO-90, if applicable, and then train a multi-task generalist policy on the same dataset based on the pretrained outcomes (e.g., encoders, skill tokens).

\textbf{Baselines.} For multitask learning, our comparison includes \algo~and three network architectures for multitask behavior cloning (BC): ResNet-RNN, ResNet-T, and ViT-T, as implemented in \citep{libero}. These architectures utilize either ResNet-18~\citep{resnet} or ViT~\cite{vit} to encode the pixel observations and then apply either a transformer or a LSTM module as temporal backbone to process a sequence of visual tokens. 
Recall that the main architecture of \algo~is the same as ResNet-T, except we add the vector quantization as well as other modules unique to \algo. 
Additionally, we compare with ACT \citep{aloha}, an IL algorithm that learns a generative model for sequences of actions, termed ``action chunks," rather than single-step actions. These action chunks can be seen as a form of temporal action abstraction. 
During policy rollout, ACT combines overlapping action chunks predicted by its policy using a weighted average. 
Thus, the critical hyperparameter of ACT is the length of the action chunk, which we set at 8 after evaluating the best option from five choices: $\{3,5,8,10,15\}$.

\textbf{Experimental Results.} Figure~\ref{fig:libero90_multitask} presents the average success rate across 90 tasks in the LIBERO-90 dataset.
It clearly demonstrates the significance of temporal action abstraction for knowledge sharing across diverse tasks, as Multitask ACT significantly outperforms the baseline BC approaches. 
More importantly, the use of pretrained skill tokens in \algo~ further leads to a substantial performance improvement over all other existing algorithms, underscoring the efficacy of \algo's pretrained skill tokens.
For the details (e.g., per-task success rateas well as the evaluation protocol), we refer readers to Appendix~\ref{a3:multitask}.

\subsection{Few-shot adaptation to unseen tasks}
In addition to learning a generalist multitask policy, we show that the learned skill token of \algo can also make learning a new task more efficient.  
Here, we evaluate the 5-shot IL performance for unseen tasks in both MetaWorld and LIBERO.

\textbf{Baselines.} We compare \algo~with the baselines below.
(1) BC from scratch. 
(2) BC with pretrained CNN encoder from \algo: This is a more computationally efficient version of the first. We find that whether freezing the CNN encoder or not does not have statistically significant effects on the downstream performance. 
(3) \algo~without BPE: This is the same as \algo with a skill-token vocabulary $V=\{1,...,C\}$ of just the quantized action codes. That is, there is no temporal abstraction. This baseline can be viewed not only as an ablated version of \algo~but also as an analog to existing hierarchical skill learning approaches such as OPAL~\citep{ajay2021opal} and TACO-RL~\citep{rosete2022tacorl}. We refer readers to Section~\ref{sec:main:relw} and Section~\ref{sec:appendix:relw} for a detailed discussion. 
(4)  ACT~\citep{aloha} from scratch. 
(5) ACT~\citep{aloha} with pretrained CNN encoder from~\algo. We freeze the CNN encoder and finetune other parts of the model.

In few-shot learning, we train each baseline algorithm for 30 epochs and evaluate each algorithm every 3 epochs. 
We then report the best success rate for the 10 evaluated checkpoints.

\begin{figure}[!htbp]
\centering
\includegraphics[width=0.85\columnwidth]{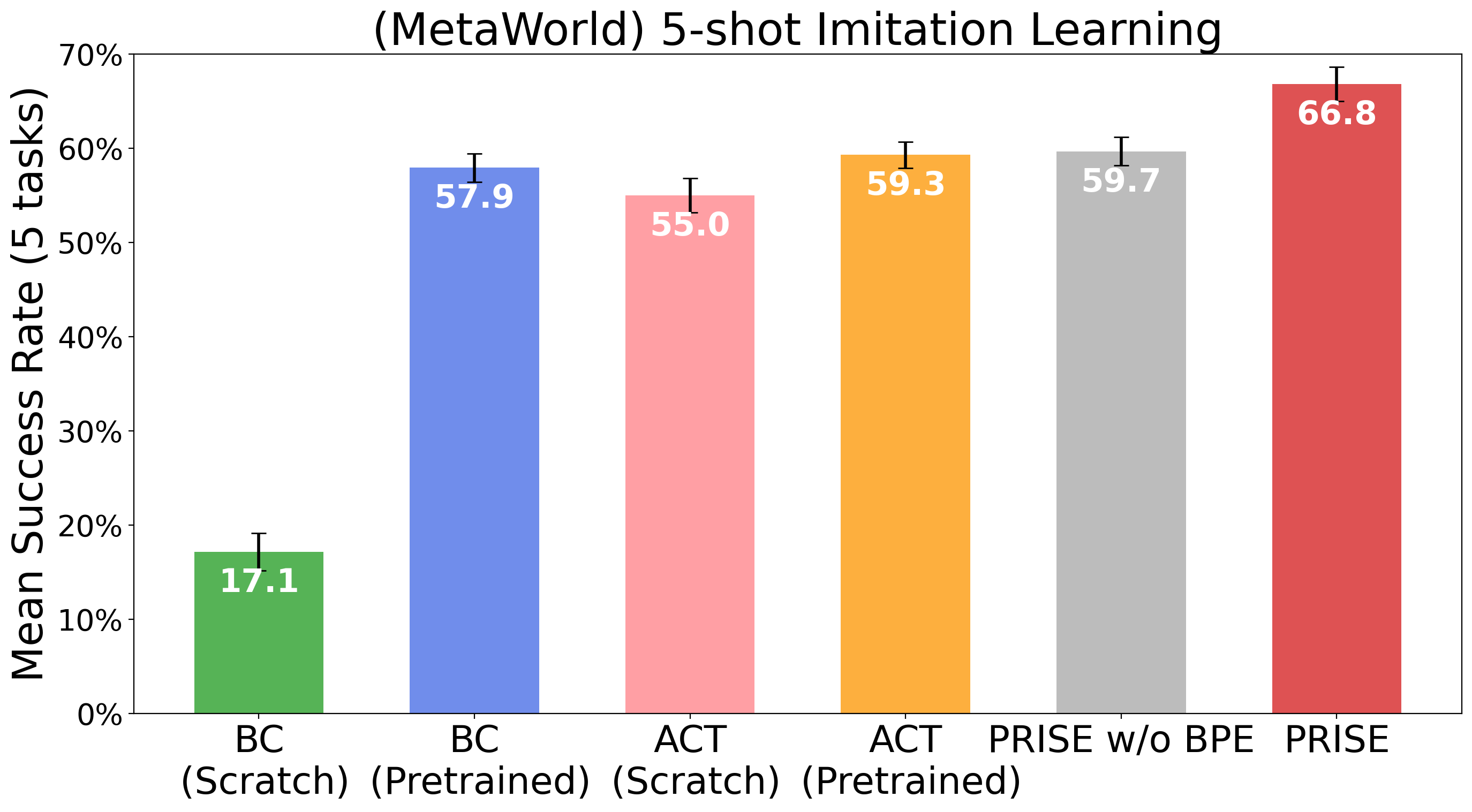}
\vspace{-0.5em}
\caption{(Metaworld): 5-shot IL performance averaged across 5 unseen tasks. Error bar represents the standard deviation across 4 random seeds.}
\vspace{-0.5em}
\label{fig:metaworld_unseen}
\end{figure}

\textbf{Experimental Results (MetaWorld).}
In ~\Cref{fig:metaworld_unseen}, we plot the averaged task success rate of 5-shot IL across 5 unseen tasks in MetaWorld. 
As shown in the figure, \algo~surpasses all other baselines (including \algo w/o BPE) with a large margin, highlighting the effectiveness of the learned temporally extended skill tokens in adapting to unseen downstream tasks. 
Furthermore, we observe that with the pretrained visual encoder from \algo, BC and ACT consistently yields improved results, indicating that the \algo~learning objective is also beneficial for pretraining visual representations from multitask offline data.

\begin{figure}[!h]
\centering
\vspace{-0.5em}
\includegraphics[width=0.85\columnwidth]{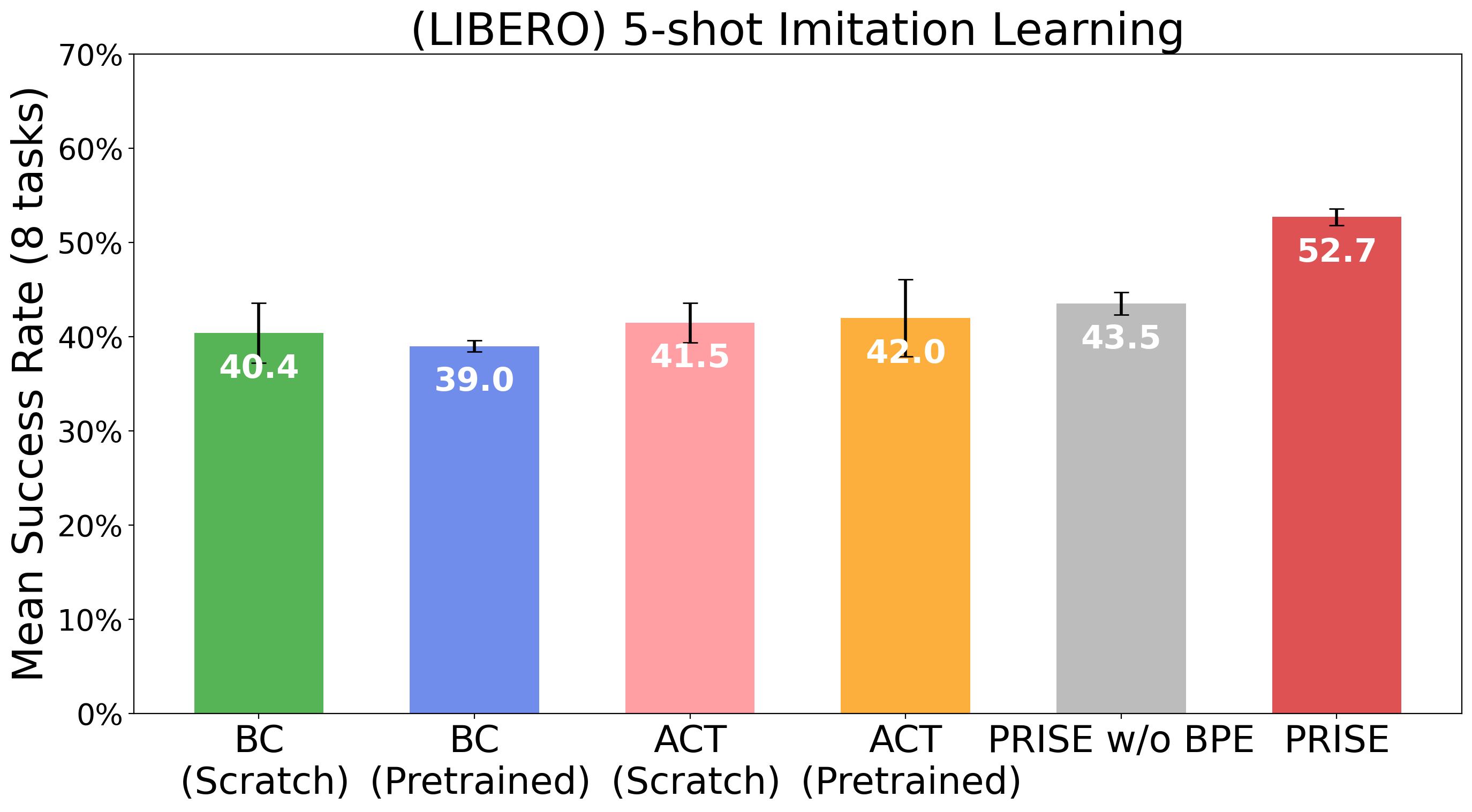}
\vspace{-1em}
\caption{(LIBERO): 5-shot IL performance averaged across 8 unseen tasks. Error bar stands for the standard deviation across 4 random seeds.}
\vspace{-1em}
\label{fig:libero_unseen}
\end{figure}

\textbf{Experimental Results (LIBERO).}
We present the average 5-shot IL performance across 8 tasks of LIBERO in~\Cref{fig:libero_unseen}.
Different from MetaWorld, as shown in the figure, \algo~pretrained encoder does not improve the performance of the base IL algorithms for BC and ACT. 
This is consistent with what is reported in~\citep{libero}, where they observe that pretraining on LIBERO-90 with multitask BC even leads to negative transfer to downstream tasks. 
However, with pretrained skill tokens from~\algo, we could improve the average success rate by 9.2\% compared with the best baseline algorithms. 
This further demonstrates the effectiveness of the proposed skill tokenization mechanisms by~\algo.

\subsection{Ablation Analyses}
\begin{figure}[!h]
\centering
\vspace{-0.5em}
\includegraphics[width=0.95\columnwidth]{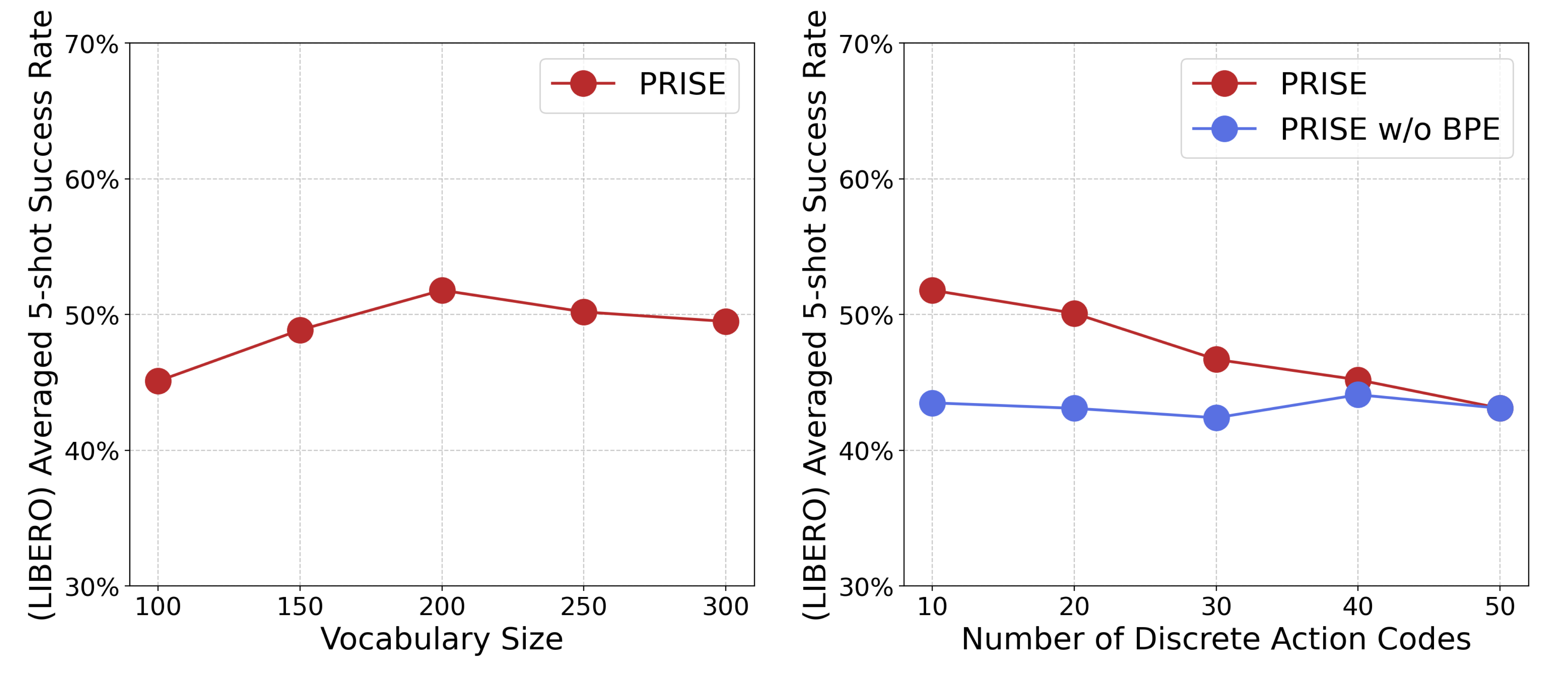}
\vspace{-1em}
\caption{(\textbf{Left}) Mean success rate of \algo~ across 8 unseen LIBERO tasks with different vocabulary size. (\textbf{Right}) The mean success rate of \algo~across 8 unseen LIBERO tasks with a different numbers of discrete action codes pre-trained.}
\vspace{-1.0em}
\label{fig:vocab_code_size}
\end{figure}

\textbf{Vocabulary Size.}
The vocabulary size of the BPE tokenizer, $|\mathcal{V}|$, plays an important role in \algo. 
A larger vocabulary size captures more motion patterns but increases the complexity of learning the skill-token policy due to more prediction classes. In addition, it learns less frequent tokens, which may not generalize or useful to downstream tasks.
On the other hand, a smaller vocabulary size limits the range of motion patterns that can be captured, but may reduce overfitting (see \cref{sec:discussion of HPs}).
In the previous experiments, we set $|\mathcal{V}|$ to be 200.
Here, we evaluate five hyperparameter settings $\{100, 150, 200, 250, 300\}$ for $|\mathcal{V}|$ to assess their impact on downstream policy adaptation, in terms of the averaged 5-shot IL performance across 8 unseen LIBERO tasks.
As \Cref{fig:vocab_code_size} (left) illustrates, \algo's performance remains relatively stable across different hyperparameter settings, except when $|\mathcal{V}|$ is extremely large or small.
This demonstrates the robustness of \algo~to the choice of $|\mathcal{V}|$, indicating that as long as the vocabulary size is not too large or small, the performance should not vary significantly.

\begin{figure}[!h]
\centering
\vspace{-0.5em}
\includegraphics[width=0.95\columnwidth]{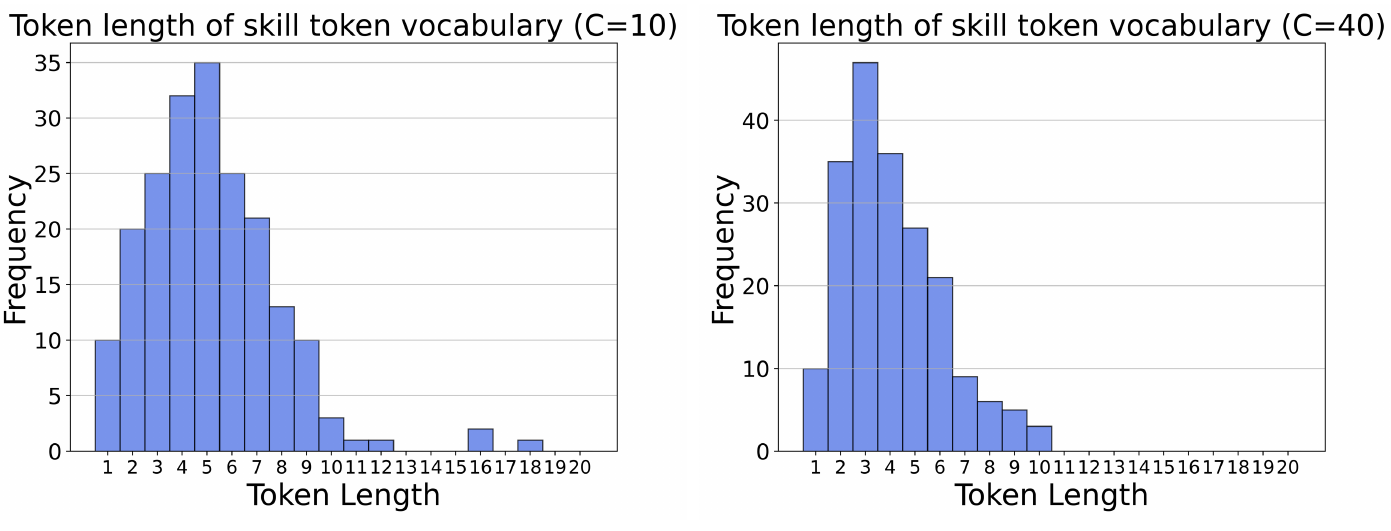}
\vspace{-1em}
\caption{(\textbf{Left}) Histogram of token length when $C=10$. (\textbf{Right}) Histogram of token length when $C=40$.}
\vspace{-0.5em}
\label{fig:hist}
\end{figure}

\textbf{Number of Action Codes.} Additionally, we verify the effects of varying the number of action codes, $C$. As $C$ increases, action patterns become less distinct, making it challenging for the BPE tokenization algorithm to identify long skill tokens common across tasks. This diminishes the advantages of temporal action abstraction. In previous experiments, $C$ was set to 10. We now present an analysis of the averaged 5-shot IL performance in LIBERO (consistent with the previous ablation) across five code size options ${10,20,30,40,50}$, while maintaining a constant vocabulary size of $|\mathcal{V}|=200$. 
As shown in \Cref{fig:vocab_code_size} (right), it is evident that with larger code sizes, such as $C=40$ or $C=50$, the performance discrepancy between \algo and \algo without BPE narrows. Furthermore, we also compare the histogram of skill token lengths for $C=10$ versus $C=40$ in \Cref{fig:hist} . The comparison reveals a clear reduction in token length with an increase in the number of codes. \looseness=-1

\begin{figure}[!h]
\centering
\includegraphics[width=0.65\columnwidth]{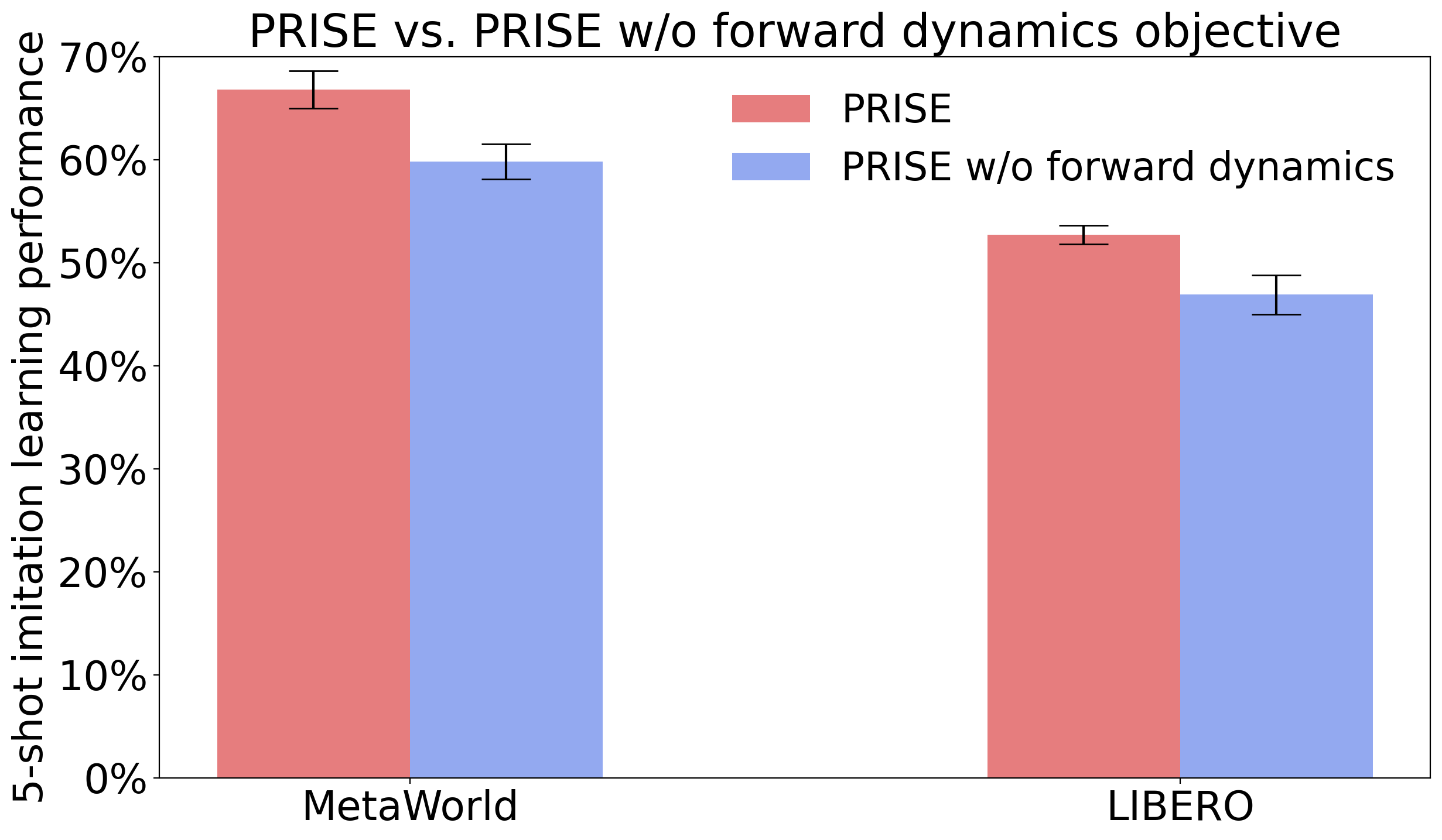}
 \vspace{-1em}
\caption{Performance of \algo with and without the forward dynamics objective on Metaworld and LIBERO.}
\vspace{-1.0em}
\label{fig:nospr}
\end{figure}

\textbf{Latent-forward-dynamics objective is crucial for learning good action codes.}
The forward-dynamics objective introduced in \algo~plays a crucial role in learning action code. 
Since we aim to learn a decoder  $\psi(z_t, e_t)$ depending on the latent variable $z_t$, without the forward prediction term, the decoder $\psi(z_t, e_t)$ may learn to depend solely on $z_t$ to reconstruct the action $a_t$, ignoring the action code $e_t$ and leading to the collapse of action codes (i.e., different codes being decoded into identical actions).
In Figure~\ref{fig:nospr}, we empirically compare the few-shot IL performance of~\algo in LIBERO with and without the forward dynamics objective. Indeed, we see a performance degradation in both domains when the forward dynamics objective is removed.

Additionally, we conducted an extra experiment on LIBERO to illustrate the action collapsing problem further. Here, we sample a batch of 5000 data points $\{(z_i, a_i)\}_{i=1}^{5000}$. For each state $z_i$, we calculate a matrix $\mathcal{M}^{(i)}$ of size $C\times C$ that quantifies the difference between the decoded action distributions of each pair of codes. 
Specifically, for entry $(j,k)$ of $\mathcal{M}$, we compute the Monte Carlo estimate of the KL divergence $\mathcal{D}_{\text{KL}}(\psi(\cdot|z_i, e_j), \psi(\cdot|z_i, e_k))$ with 1000 samples. 
Then, we compute $\mathcal{M}$ as the average of $\mathcal{M}^{(i)}$. If the action distributions of different codes significantly differ, the non-diagonal entries of $\mathcal{M}$ should be large. Conversely, if some codes are decoded into similar action distributions, we expect to see small non-diagonal entries in $\mathcal{M}$. We compare the value $\zeta=\displaystyle \frac{\text{sum}(\mathcal{M})}{\binom{C}{2}}$ for \algo with and without the forward dynamics objective.
\vspace{-1.5em}
\begin{table}[!h]
\vskip 0.15in
\begin{center}
\begin{small}
\begin{sc}
\resizebox{0.7\columnwidth}{!}{%
\begin{tabular}{lcc}
\toprule
{} & \algo & \algo w/o forward dynamics \\
\midrule
$\zeta$ & 72.42 & 14.53 \\
\bottomrule
\end{tabular}}
\end{sc}
\end{small}
\end{center}
\vskip -0.1in
\end{table}
As illustrated in the table, \algo without the forward dynamics objective exhibits a significantly smaller code-wise distance, demonstrating the necessity of the latent-forward-dynamics objective.

\begin{figure}[!h]
\centering
\includegraphics[width=0.5\columnwidth]{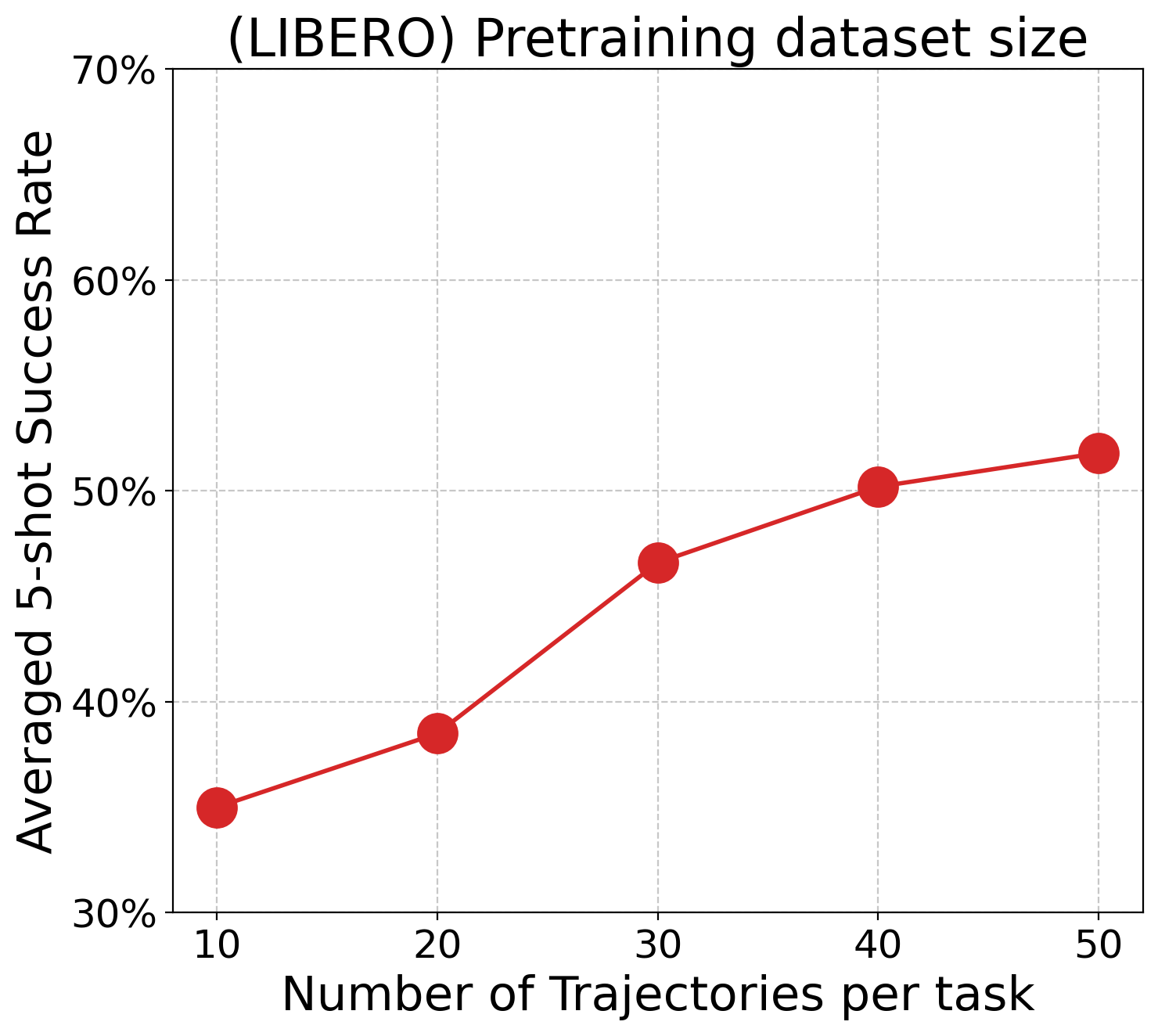}
\vspace{-1.0em}
\caption{Downstream performance of \algo under different pretraining dataset sizes}
\label{fig:dataset_size}
\vspace{-1.0em}
\end{figure}

\textbf{Effects of Dataset Size.}
To further study how much data is needed to learn a good temporal action abstraction in \algo, we conduct a set of additional experiments on LIBERO. 
Instead of using the entire LIBERO-90 dataset, where each task includes 50 human demonstration trajectories, we now sample $N$ trajectories per task, with $N\in \{10,20,30,40,50\}$. We then evaluate the downstream performance of the \algo skill token. The result is shown in Figure~\ref{fig:dataset_size}.
\vspace{-0.3em}

From the Figure, it is evident that the performance of \algo monotonically gets better as the number of pretraining trajectories per task increases. Notably, the performance of \algo is good with up to 40 trajectories per task but drops significantly with fewer than 30 trajectories per task. It would be interesting for future work to further scale up the existing LIBERO-90 dataset so that \algo may further benefit from an even larger pretraining dataset. But this should be beyond the scope of the current work.

\section{Conclusion and Future Work}
In this work, we propose a new temporal action method for continuous control scenarios, ~\algo, that leverages powerful NLP methodology during pretraining for sequential decision making tasks.
By first pretraining a vector quantization module to discretize the action space into codes and then applying \bpe to learn temporally extended skill tokens, \algo captures diverse motion patterns across pretraining tasks, enabling efficient multitask policy learning and few-shot adaptation to unseen problems.
One exciting future direction is to further scale up this approach to large real-robot datasets with diverse embodiments, such as Open X-Embodiment~\cite{rtx}. 
Additionally, instead of finetuning the model to different downstream tasks tabula rasa, we could leverage the pretrained tokens to instruction-finetune an existing large language model, so that we could capitalize on its generalization power across different tasks and scenarios.

\section*{Impact Statement}
This paper presents a pretraining algorithm to advance policy learning for sequential decision making problems. The proposed method is demonstrated in experiments of simulation robotic manipulation. As it is, this work does not have significant societal consequences. However, when the method here is scaled up to train large, foundation models, it would build AI/robotics systems that can more efficiently adapt and generalize. Such an improvement may pose potential societal risks (such as taking over human labour).

\section*{Acknowledgements} 
Ruijie Zheng and Furong Huang are supported by National Science Foundation NSF-IIS-2147276 FAI, DOD-ONR-Office of Naval Research under award number N00014-22-1-2335, DOD-AFOSR-Air Force Office of Scientific Research under award number FA9550-23-1-0048, DOD-DARPA-Defense Advanced Research Projects Agency Guaranteeing AI Robustness against Deception (GARD) HR00112020007, Adobe, Capital One and JP Morgan faculty fellowships.

\bibliography{references}
\bibliographystyle{icml2024}

\newpage
\appendix
\onecolumn
\section{Detailed Related Work \label{sec:appendix:relw}}

\textbf{Why learn temporal action abstractions?} Learning temporally extended action primitives has a long history in sequential decision-making literature. In fully observable and partially observable MDPs~\citep{puterman94,sutton2018}, the standard mathematical model for decision-making, every action takes one time step to execute. In most RL, IL, and planning algorithms, the decision space consists of individual actions, so standard methods produce policies that make a separate decision at every time step. It has long been informally observed and  shown formally under various assumptions that the difficulty of policy learning scales with the problem horizon~\citep{ross2011reduction,jiang2015,rajaraman2020toward,xie2022bellmanconsistent}. This motivates hierarchical  approaches~\citep{parr98rl,barto03advances,le2018hierarchical,nachum2018dataefficient,kipf2019compile,kujanpaa23hierquantized, jiang2022learning}, which view the decision-making process as consisting of a high-level policy that identifies task substeps and invokes lower-level temporally extended routines sometimes called \emph{options}~\citep{sutton99options} or \emph{primitives}~\citep{ajay2021opal}
to complete them. Conceptually, our \algo\ method can be viewed as hierarchical imitation learning (HIL).

\textbf{Temporal abstraction discovery for decision-making.} 
Recent methods that learn temporally extended action primitives and subsequently use them for shortening the effective decision-making horizon during high-level policy induction include CompILE~\citep{kipf2019compile}, RPL~\citep{gupta20relay}, OPAL~\citep{ajay2021opal}, TAP~\citep{jiang2023efficient}, and ACT~\citep{aloha}. 
They operate in two stages, learning the primitives during the first and applying them to solve a downstream task during the second, possibly adapting the primitives in the process. It is during this latter stage that the learned primitives provide their temporal abstraction benefits. This is subtly but crucially different from methods like DDCO~\citep{krishnan17ddco}, which learn the primitives and a higher-level policy simultaneously for a given task and benefit from the primitives via non-temporal mechanisms. 

\algo's use of BPE is dissimilar from any other method of inducing temporal action abstractions that we are aware of, and sidesteps some of the challenges of these other methods. E.g., CompILE, which, like \algo, learns \emph{variable}-duration primitives,  does so by segmenting pretraining demonstrations into semantically meaningful subtasks, for which it needs to know the number of subtasks \emph{in each pretraining trajectory} and is sensitive to errors in these values~\citep{kipf2019compile}. RPL and OPAL avoid this complication but learn primitives of a fixed duration, determined by a hyperparameter. This hyperparameter is nontrivial to tune, because short primitive durations yield little gains from temporal abstraction during learning, and long ones cause many primitives to idle after achieving a subtask. This is distinct from \algo's hyperparameter $K$, which merely puts an \emph{upper bound} on skills' duration.

We view ACT~\citep{zhao2023learning} as the most comparable method to \algo and use it as a baseline in our experiments. Like \algo, ACT handles high-dimensional pixel observations out of the box and, crucially, uses BC during downstream learning. In contrast CompILE, RPL, and OPAL assume access to ground-truth states and rely on online RL for finetuning. In many physical continuous control scenarios such as robotics, BC is arguably a more practical algorithm, and benefits that temporal abstractions provide for BC are expected to be different from those in RL, where they enable more efficient exploration.

\textbf{Policy quantization methods.} \algo's state-conditioned policy quantization method is related to SAQ~\citep{luo2023actionquantized}, but, crucially, also works for pixel observations characteristic of many continuous control scenarios in robotics. The policy encoder at the heart of \algo\ and SAQ is   VQ-VAE~\citep{oord2018neural}. Conditional VAE~\citep{kingma2014vae} is another common choice for this role (see, e.g., \citet{lynch2019learning} and \citet{kipf2019compile}).  

\textbf{Temporal abstractions versus skills in robot learning.}  
Many works on robot learning focus on the acquisition of \emph{skills}. While seemingly similar to an option or a primitive, this term has a somewhat different meaning, and most of these works don't use skills for temporal abstraction, with TACO-RL~\citep{rosete2022tacorl} being a notable exception. Namely, approaches such as DMP ~\citep{pastor09skills}, ~\citep{hausman18skills}, Play-LMP~\citep{lynch2019learning}, and MCIL~\citep{mcil21rss} aim to learn a multi-task goal-conditioned policy, and understand a skill as a latent plan for achieving a goal or several goal variations starting from a variety of initial states. In this paradigm, learning a multitask policy constitutes embedding plans into a continuous latent space, which happens without using those plans to shorten a task's horizon. Indeed, the learned skills usually don't have a termination condition (although TACO-RL assumes having a mechanism for detecting subgoal attainment), and policies produced by the aforementioned methods resample a latent skill at every time step. This is in contrast to \algo\ and, e.g., TAP~\citep{jiang2023efficient} that learn temporally extended action primitives and apply them for several time steps during deployment.

\textbf{Pretraining data assumptions.} In terms of the assumptions \algo\ imposes on the trajectories it uses for pretraining, it is more similar to C-BeT~\cite{cui2023cbet}, Play-LMP~\citep{lynch2019learning}, RPL~\citep{gupta20relay}, MCIL~\citep{mcil21rss}, and TACO-RL~\citep{rosete2022tacorl} rather than to IL or offline RL. 
Namely, \algo\ needs this data to contain meaningful behavioral patters, which it extracts using an imitation-like procedure (as does RPL). While \algo\ could be applied to arbitrary-quality data commonly assumed by offline RL or some state representation pretraining algorithms such as ~\cite{schwarzer2021pretraining, sun2023smart, wei2023imitation, zheng2023taco, zheng2024premiertaco, nicklas22tdmpc, neurips2022yuan}, many of the patterns there are unlikely to be useful, and may not easily aggregate into common temporally extended sequences that BPE aims to extract. On the other hand, the pretraining data doesn't need to be imitation-quality either.

\textbf{Tokenization in language models}
Subword tokenization methods such as BPE~\citep{gage94bpe}, Unigram~\citep{kudo2018subword}, and WordPiece~\citep{devlin2018bert} play an important role in training modern large language models. These algorithms learn tokens from a large corpus texts to mine reusable patterns. A trained tokenizer is used to compress training text data into tokens, and a large language model is trained to predict over this compressed tokens. Compared to directly predicting at the alphabet level, predicting tokens (with a proper vocabulary size) allows better generalization, because the effective sequence length that the model needs to predict is shorter.  

Next token prediction in NLP is a special case of behavior cloning in the context of decision making, where the state is the history of tokens (context).
Therefore, our algorithm can be viewed as extension of the next-token-prediction paradigm in NLP to the continuous control domain.
However, our setup introduces additional complexities not faced in NLP. 
Our raw action space is continuous rather than discrete finite alphabets in NLP. In addition, our decision agents need to consider high-dimensional image observations as part of the state representation, whereas in NLP the history is simply past tokens.
Namely, if we view alphabets in NLP as action vectors here, the BC problem in NLP has a stateless dynamical system, but in, e.g., robotics it has a non-trivial internal state. 
These complications motivate the need for observation and action encodes as well as an action quantizer.

\section{Additional Experimental Results}
\label{a1:addtional_results}
In Figure~\ref{fig:metaworld_unseen}, we present the results of mean success rate across 5 tasks in MetaWorld. Here we present the detailed results for each of the downstream unseen task.
\begin{table}[!h]\small
\centering
\renewcommand{\arraystretch}{1.4}
\resizebox{0.8\columnwidth}{!}{%
\setlength{\tabcolsep}{3pt}
\begin{tabular}{ p{2.3cm}<{\centering} p{1.8cm}<{\centering} p{1.8cm}<{\centering} p{1.8cm}<{\centering} p{1.8cm}<{\centering}  p{1.8cm}<{\centering} >{\columncolor{gray!25}}p{2.3cm}<{\centering}}
\toprule
\textbf{MetaWorld}  &   \multicolumn{6}{c}{Algorithms}   \\ 
   \textbf{Unseen Tasks} & \makecell{BC \\ (Scratch)} & \makecell{BC \\ (Pretrained)} & \makecell{ACT \\ (Scratch)} & \makecell{ACT \\ (Pretrained)} &  \makecell{\algo~w/o \\ BPE}  & \textbf{\algo}\\  
\hline
Hand Insert &  $18.8 \pm 12.5$ & $46.3\pm 7.6$ &   $45.6 \pm 6.1$    &  $56.3\pm 2.8$    &  $54.2\pm 3.9$ &   $\bm{60.0\pm 5.0}$  \\
\hline
Box Close &  $19.4 \pm 8.7$ & $53.9\pm 4.2$ &   $55.6 \pm 6.7$    &  $58.1\pm 6.8$    &  $54.8\pm 5.1$ &   $\bm{60.8\pm 6.6}$  \\
\hline
Disassemble &  $8.1 \pm 17.3$ & $65.6\pm 3.6$ &   $65.2 \pm 2.9$    &  $51.8\pm 8.5$    &  $69.6\pm 3.2$ &   $\bm{74.1\pm 7.3}$  \\
\hline
Stick Pull &  $20.6 \pm 5.8$ & $56.3\pm 5.6$ &   $51.8 \pm 4.3$    &  $65.0\pm 4.5$    &  $59.3\pm 6.2$ &   $\bm{67.5\pm 5.6}$  \\
\hline
Pick Place Wall &  $19.4 \pm 6.4$ & $67.5\pm 3.7$ &   $56.9 \pm 5.9$    &  $65.2\pm 6.5$  &  $62.5\pm 4.8$ & $\bm{71.7\pm 5.7}$  \\
\hline
\textbf{Mean} & $17.3\pm 4.2$ & $57.9\pm 1.4$ &  $55.0\pm 1.8$   &    $59.3\pm 2.0$  & $60.1\pm 1.5$ & $\bm{66.8\pm 1.8}$ \\
\bottomrule
\end{tabular}}
\caption{5-shot imitation learning performance on five unseen tasks in MetaWorld. Results are averaged across 4 random seeds.}
\label{tab:mw_unseen_task}
\vspace{-1em}
\end{table}

For LIBERO, in Table~\ref{tab:libero_task_name}, we first provide the task instruction for each of the downstream tasks.
\begin{table}[!h]
\vskip 0.15in
\begin{center}
\begin{small}
\begin{sc}
\resizebox{1.0\columnwidth}{!}{%
\begin{tabular}{lcc}
\toprule
Task ID  & Task Scene & Task Instruction \\
\midrule
0 & living room scene2 & put both the alphabet soup and the tomato sauce in the basket \\
1 & living room scene2 & put both the cream cheese box and the butter in the basket \\
2 & kitchen scene3 & turn on the stove and put the moka pot on it \\
3 & kitchen scene4 & put the black bowl in the bottom drawer of the cabinet and close it \\
4 & living room scene5 & put the white mug on the left plate and put the yellow and white mug on the right plate \\
5 & study scene1 & pick up the book and place it in the back compartment of the caddy \\
6 & living room scene6 & put the white mug on the plate and put the chocolate pudding to the right of the plate \\
7 & living room scene1 & put both the alphabet soup and the cream cheese box in the basket \\
\bottomrule
\end{tabular}}
\caption{Language instructions for 8 LIBERO downstream tasks.}
\label{tab:libero_task_name}
\end{sc}
\end{small}
\end{center}
\vskip -0.1in
\end{table}

\newpage
The detailed results for each of the LIBERO downstream unseen task are presented in Table~\ref{tab:libero_unseen_task}.
\begin{table}[!h]\small
\centering
\renewcommand{\arraystretch}{1.4}
\resizebox{0.8\columnwidth}{!}{%
\setlength{\tabcolsep}{3pt}

\begin{tabular}{ p{2.3cm}<{\centering} p{1.8cm}<{\centering} p{1.8cm}<{\centering} p{1.8cm}<{\centering} p{1.8cm}<{\centering}  p{1.8cm}<{\centering} >{\columncolor{gray!25}}p{2.3cm}<{\centering}}
\toprule
\textbf{LIBERO}  &   \multicolumn{6}{c}{Algorithms}   \\ 
   \textbf{Unseen Tasks} & \makecell{BC \\ (Scratch)} & \makecell{BC \\ (Pretrained)} & \makecell{ACT \\ (Scratch)} & \makecell{ACT \\ (Pretrained)} &  \makecell{\algo~w/o \\ BPE}  & \textbf{\algo}\\  
\hline
Task 0 &  $16.7 \pm 6.2$ & $6.7\pm 2.3$ &   $21.7\pm 8.4$    &  $20.0\pm 6.0$    &  $16.7\pm 6.2$ &   $\bm{26.7\pm 6.4}$  \\
\hline
Task 1 &  $25.0 \pm 4.7$ & $48.3\pm 10.3$ &   $26.7\pm 10.5$    & $33.3 \pm 13.1$    &  $25.0\pm 4.7$ &   $\bm{48.3\pm 9.4}$  \\
\hline
Task 2 &  $68.3 \pm 7.4$ & $60.0\pm 4.1$ &   $61.7\pm 9.2$    &  $67.7\pm 6.2$    &  $68.3\pm 6.2$ &   $\bm{70.0\pm 0.0}$  \\
\hline
Task 3 &  $66.7 \pm 7.9$ & $66.7\pm 8.4$ &   $73.3 \pm 9.3$    &  $70.3\pm 6.2$    &  $66.7\pm 5.9$ &   $\bm{78.3\pm 8.8}$  \\
\hline
Task 4 &  $25.0 \pm 10.8$ & $26.7\pm 3.1$ &   $45.0 \pm 4.3$    &  $35.0\pm 4.1$    &  $\bm{45.0\pm 10.8}$ &   $28.3\pm 6.2$  \\
\hline
Task 5 &  $70.0 \pm 0.0$ & $46.7\pm 13.1$ &   $58.3 \pm 4.3$    &  $68.3\pm 6.5$    &  $75.0\pm 4.2$ &   $\bm{90.0\pm 4.1}$  \\
\hline
Task 6 &  $23.3 \pm 4.3$ & $21.7\pm 2.4$ &   $\bm{35.0 \pm 4.3}$    &  $15.0\pm 0.0$    &  $23.3\pm 6.2$ &   $25.0\pm 4.1$  \\
\hline
Task 7 &  $28.3 \pm 4.3$ & $35.0\pm 7.1$ &   $10.0 \pm 0.0$    &  $26.7\pm 7.0$    &  $28.3\pm 10.2$ &   $\bm{45.0\pm 8.1}$  \\
\hline
\textbf{Mean} & $40.4\pm 1.2$ & $39.0\pm 4.1$ &  $41.5\pm 2.1$   &    $42.0\pm 3.2$  &    $43.5\pm 0.6$ &    $\bm{52.7\pm 0.9}$ \\
\bottomrule

\end{tabular}}
\caption{5-shot imitation learning performance across 8 unseen tasks in LIBERO. Resullts are averaged across 4 random seeds}
\label{tab:libero_unseen_task}
\vspace{-1em}
\end{table}
\onecolumn
\newpage
\section{Implementation Details}
\label{a2:implementation}

%\begin{wrapfigure}{r}{0.98\textwidth}
\begin{figure*}[!htbp]
\hspace{0.1in}
\begin{subfigure}{0.98\textwidth}
\begin{lstlisting}[language=Python]
# next_observation: A sequence of K=3 future pixel observations o_{t+1},..., o_{t+K}
# past_observation: A sequence of H historical pixel observations o_{t-H+1},..., o_{t}
# action:           A sequence of $K$ action a_t,..., a_{t+K-1}
# obs_encoder:      Observation Encoder (CNN+Transformer)
# quantizer:        Discrete action code quantizer 
# dynamics:         Latent Dynamics Model
# project:          Projection layer
# predictor:        Latent state predictor
# action_decoder:   Quantized action code decoder
# beta:             Weight of decoder loss (1.0 in MetaWorld and 0.01                       in LIBERO for numerical stability)

dynamic_losses, quantization_losses, decoder_losses = 0, 0, 0
z = state_encoder(past_observation)
z_hat = z
for k in range(K):
    z = state_encoder(past_observation)
    past_observation.push(next_observation[k])
    u_quantized, quantization_loss = quantizer(action[k], z.detach())
    quantization_losses += quantization_loss
    decode_action       =  decoder(z, u_quantized)
    decoder_losses      += action_loss(decode_action, action[k])
    z_hat  = dynamics(z_hat, u_quantized)
    z_next = state_encoder(observation[k+1]) 
    y_hat  = predictor(project(z_hat)) 
    y_next = project(z_next).detach()
    dynamic_losses += -cosine_similarity(y_hat, y_next)
(dynamic_losses + quantization_losses + beta*decoder_losses).backward() 
\end{lstlisting}
\end{subfigure}
\caption{Pseudocode for the pretraining stage I of \algo}
\label{fig:pseudocode}
\end{figure*}
%\end{wrapfigure}

\textbf{Note}: Here our decision to adapt a BYOL-like objective for the forward dynamics losses, rather than directly modeling the forward transition dynamics as in model-based learning approaches~\cite{janner2019mbpo, zheng2023is,hansen2022temporal,hafner2023mastering, wang2023coplanner}, is mainly motivated by the effectiveness of the BYOL~\cite{schwarzer2021pretraining} objective in preventing representation collapse during latent forward dynamics training, as we are jointly learning both the action quantization and state representation here in the pretraining stage I.

\textbf{Metaworld}: We generate 100 expert trajectories for each pretraining task using the scripted policy provided in MetaWorld.
The observation for each step includes an $84\times 84$ third-person view image as well as an 8-dimensional proprioceptive state vector of the robot's end-effector.  
We use the same shallow CNN as in~\cite{yarats2022drqv2} to encode the observations into a 64-dimensional latent vector and apply a linear layer to embed the 8-dimensional state vector also into a 64-dimensional latent vector. 
The architecture of the shallow CNN is presented in~\Cref{fig:shallowcnn}.
\begin{figure*}[h!]
\hspace{0.1in}
\begin{subfigure}{0.98\textwidth}
\begin{lstlisting}[language=Python]
class Encoder(nn.Module):
    def __init__(self):
        super().__init__()
        self.repr_dim = 32 * 35 * 35
        self.convnet = nn.Sequential(nn.Conv2d(84, 32, 3, stride=2),
                        nn.ReLU(), nn.Conv2d(32, 32, 3, stride=1),
                        nn.ReLU(), nn.Conv2d(32, 32, 3, stride=1),
                        nn.ReLU(), nn.Conv2d(32, 32, 3, stride=1),
                        nn.ReLU(),
                        nn.Linear(self.repr_dim, feature_dim),
                        nn.LayerNorm(feature_dim), nn.Tanh())
        self.trunk = nn.Sequential(nn.Linear(self.repr_dim, feature_dim),
                        nn.LayerNorm(feature_dim), nn.Tanh())

    def forward(self, obs):
        obs = obs / 255.0 - 0.5
        h = self.convnet(obs).view(h.shape[0], -1)
        return self.trunk(h)
\end{lstlisting}
\end{subfigure}
\caption{Architecture of the Shallow CNN encoder used in MetaWorld.}
\label{fig:shallowcnn}
\end{figure*}
Next, we apply a transformer decoder module with 4 layers and number of heads equal to 8 to extract the observation embedding. 
We set the context length to be 10.
The action decoder $\psi$ is a three-layer MLP with hidden size being 1024.

\textbf{LIBERO}: For LIBERO, we pretrain \algo~on the provided LIBERO-90 dataset.
The pretraining dataset contains 50 demonstration trajectories for each task, collected by human teleoperation. 
For each timestep, the agent observes a third-person view image, a first-person view image from its wrist camera (both with resolution $128\times 128$), a 9-dimensional vector of the robot's proprioceptive state, and a task instruction in natural language. 
We use the exact same architecture as ResNet-T in~\citep{libero}.
In particular, we use two ResNet18 encoders~\cite{resnet} to process the image observations, utilizing FiLM~\cite{film} encoding to incorporate the 768-dimensional BERT embedding of the task language instruction.
The action decoder $\psi$ is also a three-layer MLP with hidden size being 1024.

\textbf{Evaluation of Few-shot Imitation Learning} A batch size of 128 and a learning rate of 1e-4 are used for Metaworld, and a batch size of 64 is used for LIBERO. 
In total, we take 30,000 gradient steps and conduct evaluations for every 3000 steps. 
For both MetaWorld and LIBERO, we execute 40 episodes and calculate the success rate of the trained policy.
We report the highest success rates across 10 evaluated checkpoints.

\textbf{Computational Resources} For our experiments, we use 8 NVIDIA RTX A6000 with PyTorch Distributed DataParallel for pretraining \algo, and we use NVIDIA RTX2080Ti for downstream imitation learning on Metaworld, and RTX A5000 on LIBERO.

\newpage
\onecolumn
\section{Additional Results on Multitask Learning}
\label{a3:multitask}
Below we present the per-task success rate for LIBERO-90. 
The results of BC (ResNet-RNN), BC (ResNet-T), BC (ViT-T) are taken from~\citep{libero}. 
For the evaluation of ACT and~\algo, we follow the same evaluation protocol.
We take the last training checkpoint of them respectively and evaluate the success rate on each task with 20 rollouts.
\begin{table}[!h]\small
\centering
\renewcommand{\arraystretch}{1.4}
\resizebox{0.68\columnwidth}{!}{%
\setlength{\tabcolsep}{3pt}
\begin{tabular}{ p{2.3cm}<{\centering} p{2.0cm}<{\centering} p{2.0cm}<{\centering} p{1.8cm}<{\centering} p{2.0cm}<{\centering}  p{2.0cm}<{\centering} p{2.3cm}<{\centering}}
\toprule
\textbf{LIBERO-90}  &   \multicolumn{5}{c}{Multitask Algorithms}   \\ 
   \textbf{Task ID} & \makecell{BC\\(ResNet-RNN)} & \makecell{BC\\(ResNet-T)} & \makecell{BC\\(ViT-T)} & \makecell{ACT} & \textbf{\algo}\\ 
   \midrule
   0  &  $0.45$ & $0.45$ &   $0.6$    &  $1.0$    &  $0.8$  \\
   1  &  $0.1 $ & $0.1$ &   $0.0$    &  $0.6$    &  $0.35$  \\
   2  &  $0.0 $ & $0.25$ &  $0.05$    &  $0.95$    &  $0.7$  \\
   3  &  $0.15$ & $0.0$ &   $0.0$    &  $0.4$    &  $0.5$  \\
   4  &  $0.2 $ & $0.0$ &   $0.0$    &  $0.3$    &  $0.45$  \\
   5  &  $0.25$ & $0.0$ &   $0.0$    &  $0.7$    &  $0.65$  \\
   6  &  $0.05$ & $0.0$ &   $0.15$    &  $0.4$    &  $0.5$  \\
   7  &  $0.3$ & $0.45$ &   $0.25$    &  $1.0$    &  $0.95$  \\
   8  &  $0.05$ & $0.0$ &   $0.0$    &  $1.0$    &  $0.6$  \\
   9  &  $0.35$ & $0.3$ &   $0.0$    &  $0.95$    &  $0.35$  \\
   10 &  $0.2$ & $0.7$ &    $0.15$    &  $1.0$    &  $0.95$  \\
   11 &  $0.45$ & $0.4$ &   $0.65$    &  $1.0$    &  $0.95$  \\
   12 &  $0.2$ & $0.05$ &   $0.35$    &  $0.35$    &  $0.2$  \\
   13 &  $0.1$ & $0.35$ &   $0.0$    &  $0.25$    &  $0.4$  \\
   14 &  $0.35$ & $0.1$ &   $0.0$    &  $0.75$    &  $0.35$  \\
   15 &  $0.5 $ & $0.1$ &   $0.6$    &  $0.95$    &  $0.75$  \\
   16 &  $0.25$ & $0.05$ &  $0.25$    &  $0.75$    &  $0.4$  \\
   17 &  $0.05$ & $0.05$ &  $0.0$    &  $0.5$    &  $0.15$  \\
   18 &  $0.4$ &  $0.3$ &   $0.35$    &  $0.25$    &  $0.3$  \\
   19 &  $0.15$ & $0.0$ &   $0.4$    &  $1.0$    &  $0.65$  \\
   20 &  $0.6$ &  $0.7$ &   $1.0$    &  $1.0$    &  $1.0$  \\
   21 &  $0.15$ & $0.0$ &   $0.35$    &  $0.7$    &  $0.3$  \\
   22 &  $0.75$ & $0.4$ &   $0.55$    &  $0.75$    &  $0.85$  \\
   23 &  $0.2$ &  $0.0$ &   $0.05$    &  $0.45$    &  $0.05$  \\
   24 &  $0.25$ & $0.3$ &   $0.25$    &  $0.1$    &  $0.95$  \\
   25 &  $0.8$ & $0.6$ &   $0.85$    &  $0.1$    &  $0.9$  \\
   26 &  $0.0$ & $0.0$ &   $0.35$    &  $0.6$    &  $0.55$  \\
   27 &  $0.05$ & $0.0$ &   $0.2$    &  $0.35$    &  $0.05$  \\
   28 &  $0.05$ & $0.8$ &   $0.9$    &  $1.0$    &  $1.0$  \\
   29 &  $0.2$ &  $0.1$ &   $0.15$    &  $0.85$    &  $1.0$  \\
   30 &  $0.05$ & $0.05$ &   $0.1$    &  $0.4$    &  $0.5$  \\
   31 &  $0.5$ &  $0.25$ &   $0.2$    &  $1.0$    &  $0.85$  \\
   32 &  $0.0$ & $0.0$ &   $0.05$    &  $0.3$    &  $0.2$  \\
   33 &  $0.1$ & $0.1$ &   $0.2$    &  $0.5$    &  $0.3$  \\
   34 &  $0.2$ & $0.05$ &   $0.1$    &  $0.5$    &  $0.8$  \\
   35 &  $0.7$ & $0.1$ &   $0.45$    &  $1.0$    &  $0.75$  \\
   36 &  $0.05$ & $0.0$ &   $0.05$    &  $0.05$    &  $0.25$  \\
   37 &  $0.05$ & $0.05$ &   $0.15$    &  $0.0$    &  $0.3$  \\
   38 &  $0.15$ & $0.0$ &   $0.1$    &  $0.9$    &  $0.2$  \\
   39 &  $0.2$ & $0.25$ &   $0.75$    &  $0.4$    &  $0.85$  \\
   40 &  $0.45$ & $0.15$ &  $0.05$    &  $0.9$    &  $0.5$  \\
   41 &  $0.25$ & $0.4$ &   $0.3$    &  $0.85$    &  $0.55$  \\
   42 &  $0.15$ & $0.45$ &   $0.15$    &  $0.7$    &  $0.8$  \\
   43 &  $0.1$ & $0.1$ &   $0.15$    &  $0.85$    &  $0.4$  \\
   44 &  $0.5$ & $0.4$ &   $0.95$    &  $0.75$    &  $0.85$  \\
\hline
\bottomrule
\end{tabular}}
% \vspace{-1em}
\label{tab:multitask}
\vspace{-1em}
\end{table}

\begin{table}[!t]\small
\centering
\renewcommand{\arraystretch}{1.4}
\resizebox{0.68\columnwidth}{!}{%
\setlength{\tabcolsep}{3pt}
\begin{tabular}{ p{2.3cm}<{\centering} p{2.0cm}<{\centering} p{2.0cm}<{\centering} p{1.8cm}<{\centering} p{2.0cm}<{\centering}  p{2.0cm}<{\centering} p{2.3cm}<{\centering}}
\toprule
\textbf{LIBERO-90}  &   \multicolumn{5}{c}{Multitask Algorithms}   \\ 
   \textbf{Task ID} & \makecell{BC\\(ResNet-RNN)} & \makecell{BC\\(ResNet-T)} & \makecell{BC\\(ViT-T)} & \makecell{ACT} & \textbf{\algo}\\ 
   \midrule
   45 &  $0.0$ &  $0.0$ &   $0.0$    &  $0.8$    &  $0.55$  \\
   46 &  $0.0$ &  $0.0$ &   $0.0$    &  $0.0$    &  $0.35$  \\
   47  &  $0.0$ & $0.0$ &   $0.0$    &  $0.0$    &  $0.25$  \\
   48  &  $0.0$ & $0.0$ &   $0.0$    &  $0.0$    &  $0.65$  \\
   49  &  $0.0$ & $0.0$ &   $0.0$    &  $0.0$    &  $0.65$  \\
   50  &  $0.0$ & $0.0$ &   $0.0$    &  $0.35$    &  $0.4$  \\
   51  &  $0.0$ & $0.0$ &   $0.0$    &  $0.1$    &  $0.1$  \\
   52  &  $0.0$ & $0.0$ &   $0.0$    &  $0.05$    &  $0.3$  \\
   53  &  $0.0$ & $0.0$ &   $0.0$    &  $0.05$    &  $0.6$  \\
   54  &  $0.0$ & $0.05$ &  $0.0$    &  $0.15$    &  $0.5$  \\
   55  &  $0.0$ & $0.05$ &  $0.0$    &  $0.0$    &  $0.35$  \\
   56  &  $0.0$ & $0.3$ &   $0.0$    &  $0.0$    &  $0.8$  \\
   57  &  $0.0$ & $0.25$ &  $0.0$    &  $0.0$    &  $0.5$  \\
   58  &  $0.0$ & $0.0$ &   $0.0$    &  $0.5$    &  $0.2$  \\
   59  &  $0.0$ & $0.25$ &  $0.0$    &  $0.0$    &  $0.65$  \\
   60  &  $0.0$ & $0.4$ &   $0.2$    &  $0.45$    &  $0.8$  \\
   61  &  $0.0$ & $0.2$ &   $0.25$    &  $0.05$    &  $0.85$  \\
   62  &  $0.0$ & $0.0$ &   $0.1$    &  $0.05$    &  $0.4$  \\
   63  &  $0.0$ & $0.0$ &   $0.0$    &  $0.0$    &  $0.4$  \\
   64  &  $0.0$ & $0.0$ &   $0.05$    &  $0.25$    &  $0.15$  \\
   65  &  $0.0$ & $0.0$ &   $0.05$    &  $0.05$    &  $0.15$  \\
   66  &  $0.15$& $0.0$ &   $0.0$    &  $0.45$    &  $0.3$  \\
   67  &  $0.1$ & $0.15$ &  $0.05$    &  $0.55$    &  $0.55$  \\
   68  &  $0.35$ & $0.05$ & $0.05$    &  $0.6$    &  $0.85$  \\
   69  &  $0.1$ & $0.65$ &  $0.0$    &  $0.85$    &  $0.9$  \\
   70  &  $0.05$ & $0.55$ & $0.0$    &  $0.6$    &  $0.55$  \\
   71  &  $0.0$ & $0.2$ &   $0.0$    &  $0.0$    &  $0.35$  \\
   72  &  $0.0$ & $0.2$ &   $0.0$    &  $0.30$    &  $0.6$  \\
   73  &  $0.0$ & $0.0$ &   $0.05$    &  $0.35$    &  $0.3$  \\
   74  &  $0.2$ & $0.05$ &  $0.0$    &  $0.7$    &  $0.45$  \\
   75  &  $0.0$ & $0.1$ &   $0.0$    &  $0.35$    &  $0.25$  \\
   76  &  $0.0$ & $0.3$ &   $0.35$    &  $0.1$    &  $0.65$  \\
   77  &  $0.35$ & $0.3$ &  $0.0$    &  $0.7$    &  $0.8$  \\
   78  &  $0.25$ & $0.1$ &  $0.0$    &  $0.1$    &  $0.45$  \\
   79  &  $0.25$ & $0.45$ &  $0.05$    &  $0.95$    &  $0.3$  \\
   80  &  $0.0$ & $0.2$ &   $0.0$    &  $0.45$    &  $0.3$  \\
   81  &  $0.2$ & $0.0$ &   $0.0$    &  $0.5$    &  $0.35$  \\
   82  &  $0.4$ & $0.45$ &  $0.15$    &  $0.55$    &  $0.8$  \\
   83  &  $0.0$ & $0.05$ &  $0.05$    &  $0.0$    &  $0.55$  \\
   84  &  $0.0$ & $0.2$ &   $0.05$    &  $0.15$    &  $0.75$  \\
   85  &  $0.0$ & $0.0$ &   $0.05$    &  $0.1$    &  $0.75$  \\
   86  &  $0.05$ & $0.2$ &  $0.05$    &  $0.3$    &  $0.95$  \\
   87  &  $0.2$ & $0.1$ &   $0.15$    &  $1.0$    &  $0.65$  \\
   88  &  $0.0$ & $0.25$ &   $0.05$    &  $0.85$    &  $0.55$  \\
   89  &  $0.15$ & $0.1$ &  $0.05$    &  $0.45$    &  $0.55$  \\
\hline
\bottomrule
\end{tabular}}
% \vspace{-1em}
\caption{Multitask success rate on LIBERO-90.}
\vspace{-1em}
\end{table}

\begin{figure*}[!h]
\centering
\includegraphics[width=0.5\textwidth]{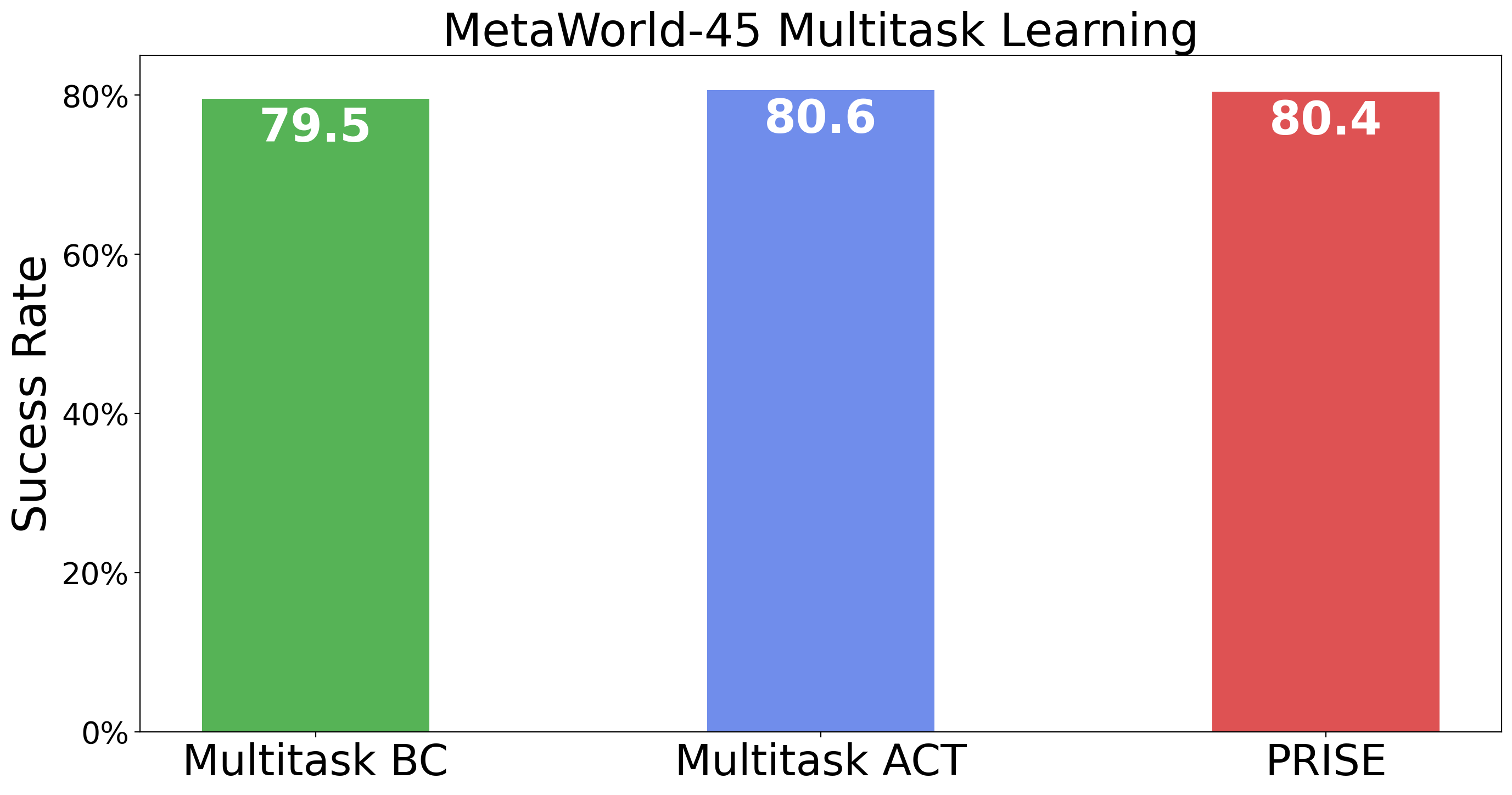}
% \vspace{-10pt}
\caption{Multitask Learning Results on MetaWorld.}
% \vspace{-15pt}
\label{fig:metaworld_mt45}
\end{figure*}

\onecolumn
\newpage
Additionally, we also provide the results of multitask performance on 45 MetaWorld pretraining tasks.
As shown in~\ref{fig:metaworld_mt45}, all three algorithms perform well on MetaWorld, with an average success rate around 80\% across 45 tasks.
Thus in this paper, we focus on the more challenging LIBERO-90 multitask benchmark.
\end{document}